\PassOptionsToPackage{table}{xcolor}
\documentclass{teleai}

\usepackage{amsmath,amssymb}
\usepackage{array,booktabs,graphicx,longtable,listings,natbib,tabularx}
\usepackage[cal=cm]{mathalpha}

\graphicspath{{assets/}{assets/supplementary/}{assets/figures/}}
\newcolumntype{Y}{>{\raggedright\arraybackslash}X}
\newcommand{\method}{WorldSearcher}
\newcommand{\evaluation}{Search-to-World}
\newif\ifdraft
\drafttrue

\newcommand{\cmark}{\textcolor{green!55!black}{$\checkmark$}}
\newcommand{\xmark}{\textcolor{red}{$\times$}}
\newcommand{\pmark}{\textcolor{orange!85!black}{$\sim$}}

\begin{document}

\title{\evaluation: Evaluation of 3D World Delivery from User Request through Web Search}

\author[1,2,3,4]{Zixiao Gu}
\author[1,\dagger]{Yabo Chen}
\author[1,5]{Xunzhi Xiang}
\author[3,4,6]{Yu He}
\author[1]{Haibin Huang}
\author[1]{Chi Zhang}
\author[2,*]{Yunbo Wang}
\author[1,*]{Xuelong Li}

\affiliation[1]{Institute of Artificial Intelligence, China Telecom (TeleAI)}
\affiliation[2]{Shanghai Jiao Tong University}
\affiliation[3]{Shanghai Innovation Institution}
\affiliation[4]{ActiMind}
\affiliation[5]{Nanjing University}
\affiliation[6]{Fudan University}

\contribution[\dagger]{Project Leader}
\contribution[*]{Corresponding Authors.}
\metadata[Keywords]{3D World Delivery; Agentic Systems; Web Search; Evaluation Benchmark; 3D Reconstruction.}
\correspondence{Yunbo Wang (\email{yunbow@sjtu.edu.cn}); Xuelong Li (\href{mailto:xuelong_li@ieee.org}{\texttt{xuelong\_li@ieee.org}})}
\date{\today}

\abstract{Agentic systems can understand user requests, search the live web, and use external tools to complete complex tasks, yet whether they can carry a user request through relevant content web retrieval to a usable 3D world has not been systematically evaluated. Both an established end-to-end pipeline and benchmark for evaluating it are missing. The core problem is bridging the gap from retrieval to delivery: relevant web-sourced visual content must be successfully turned into a request-aligned and perceptually acceptable 3D world.
We address these gaps separately. First, we introduce \evaluation, an evaluation task that defines an end-to-end pipeline from a user request through web-sourced visual content retrieval to the delivery of a usable 3D world. Observed Retrieval Rate (ORR) and World Delivery Rate (WDR) are proposed as two metrics to distinguish observing relevant web-sourced visual content from successfully delivering a 3D world. Second, we present \method, a reuse-then-reconstruction harness system that connects existing search agent systems to 3D world delivery. It first seeks available 3D worlds for reuse, and then falls back to video-based reconstruction when no reusable worlds are retrieved. We also design a structured recovery control in \method\ to revise temporal grounding, replace source videos, or reformulate queries for failure recovery. We benchmark several widely used models on the \evaluation\ task using \method\ and conduct an ablation study of supervised fine-tuning (SFT) for the core models used by recovery subagents, thereby validating the recovery control of \method. Our results reveal that \evaluation\ delivery varies with the model used by the agentic system and that observing relevant web-sourced visual content alone does not guarantee successful delivery. More importantly, joint training of the recovery agents improves both delivery success and action efficiency. Overall, \evaluation\ makes the capability of 3D world delivery by an agentic system measurable, while \method\ provides an effective harness system with complementary recovery components.
}

\maketitle

\begin{figure}[t]
  \centering
  \includegraphics[width=\linewidth]{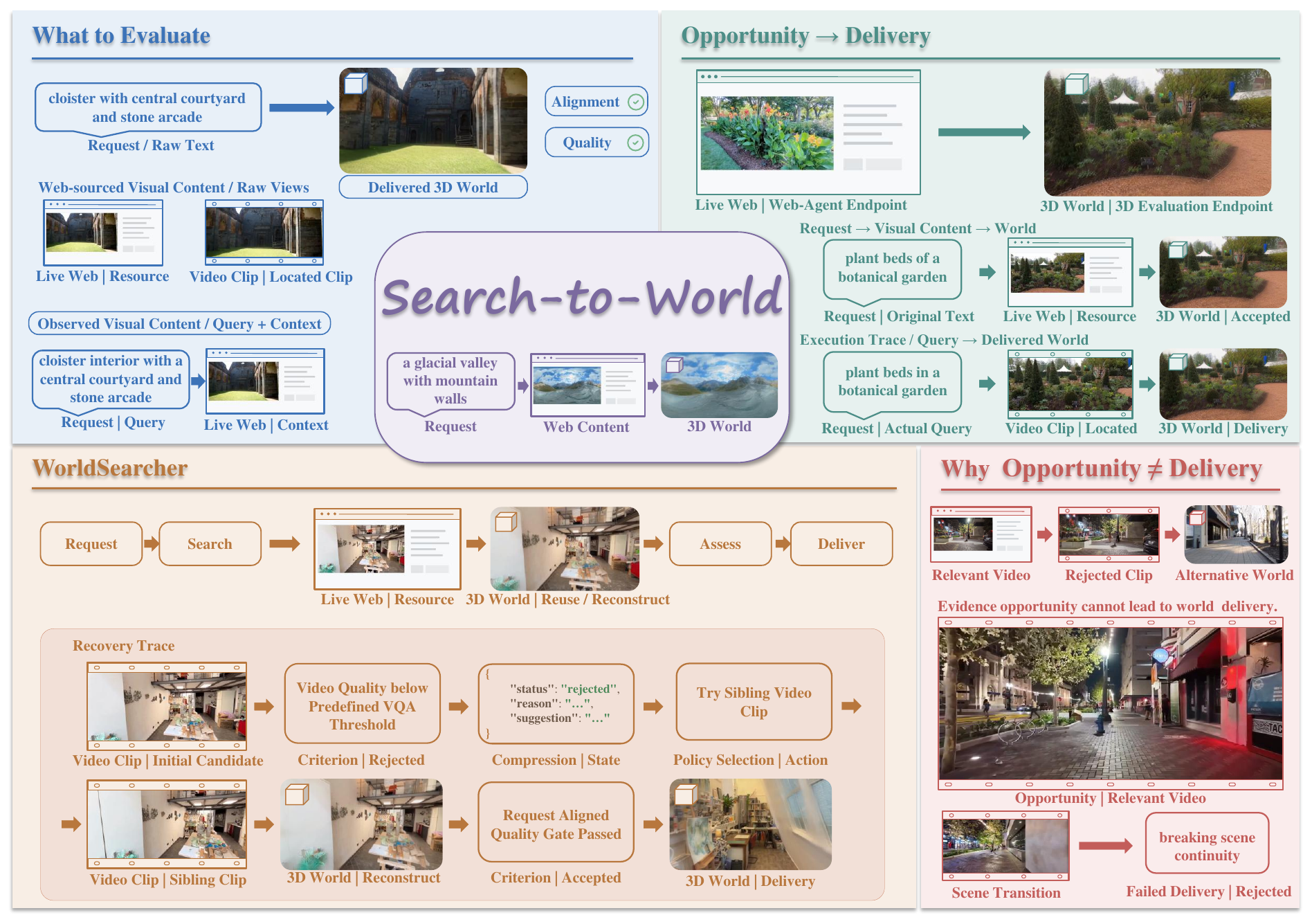}
  \caption{\textbf{\evaluation\ evaluates whether an agentic system can turn web-sourced visual content into usable 3D worlds.} The evaluation connects the user request and its observed web-sourced resources to the delivered 3D world, requiring both request alignment and perceptual quality. We evaluate not only the relevant content web retrieval, ~i.e., the ``Opportunity'', but also whether the whole system produces an acceptable world, noted as ``Delivery''. We propose \method\ which models this task as a reuse-then-reconstruction harness system with structured recovery control. When a candidate fails the delivery assessment, \method\ records a compact failure state and uses it to guide the next attempt toward an accepted world. An ``Opportunity'' is only a starting point: relevant content may lack sufficient coverage or perceptual quality for the requested world, or reuse and reconstruction may fail. By tracing each attempt to the final world, \evaluation\ identifies the path from ``Opportunity'' to ``Delivery''}
  \label{fig:new_teaser}
\end{figure}

\begin{figure}[t]
  \centering
  \includegraphics[width=\textwidth,pagebox=cropbox]{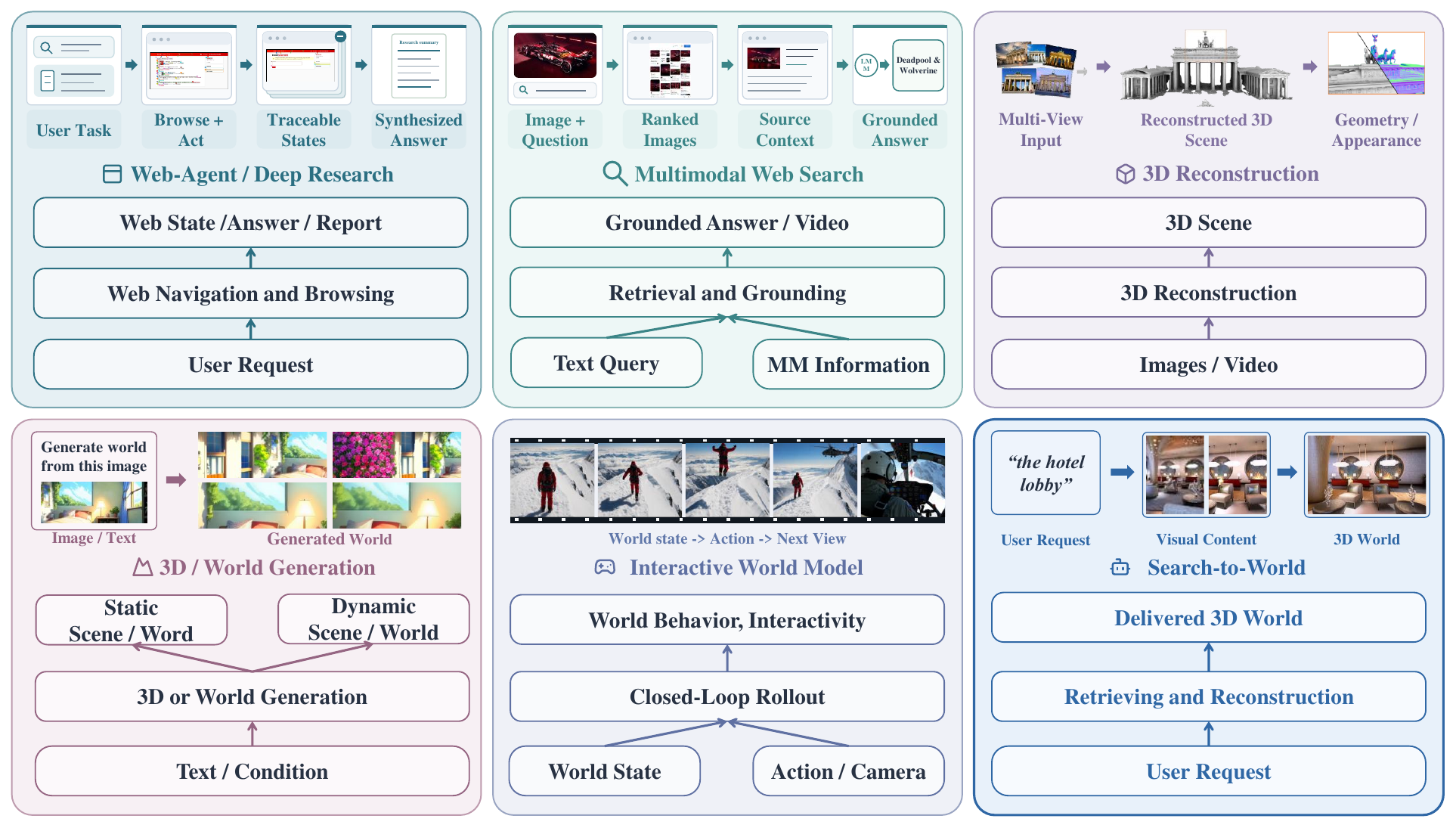}
  \caption{\textbf{Representative evaluation families use different inputs and outputs.} Web-agent and deep research benchmarks start from a user request and terminate at a web state, answer, or report. Multimodal web search benchmarks combine a text query with multimodal information as input and terminate at a grounded answer or multimodal outputs e.g. images, videos. 3D reconstruction benchmarks assume supplied images or video and terminate at a 3D scene. 3D / world generation benchmarks assess generated static or dynamic scene / worlds, while interactive world model benchmarks assess behavior and stability over closed-loop rollouts. \evaluation\ connects a user request and web-sourced visual content acquisition to the delivered 3D world, while the other families cover only parts of this end-to-end process.}
  \label{fig:evaluation_boundary}
\end{figure}

\section{Introduction}
\label{sec:introduction}

Whether an agentic system can turn a user request into a usable 3D world is an end-to-end capability question, not merely a retrieval question. As Figure~\ref{fig:new_teaser} illustrates, the requested outcome is not a web state, an answer, or relevant visual content, but a delivered 3D world that preserves request alignment and perceptual quality across the complete execution pipeline. Accomplishing this outcome requires the system to discover accessible visual content on the live web, identify the requested content, and then reuse or reconstruct a world that satisfies the request.

As summarized in Figure~\ref{fig:evaluation_boundary}, prior evaluation families cover partial stages or endpoints of the whole pipeline. Web-agent and deep-research benchmarks evaluate web interaction and answer or report production~\cite{zhou2024webarena,koh2024visualwebarena,wei2025browsecomp,yoran2024assistantbench,gou2025mind2web2,wang2026liveresearchbench}, while multimodal web-search benchmarks evaluate multimodal content acquisition and grounding~\cite{jiang2024mmsearch,tao2025mmsearchplus,du2026mmsearchplus,liang2025videobrowser,liu2026videodr,yu2026rvms}. 3D reconstruction and world-generation benchmarks evaluate scene recovery or world synthesis from supplied inputs~\cite{jensen2014largescale,schops2017multiview,knapitsch2017tanks,liang2024nvsquality,martin2025gsqa,he2023t3bench,duggal2025eval3d,tam2026sceneeval,duan2025worldscore,lu2025fourworldbench,fan2026geot2v}, while interactive-world benchmarks assume an already instantiated world and evaluate behavior within it~\cite{playworld2026,wbench2026,worldroambench2026,worldexam2026,worldolympiad2026,harnessevalw2026}. Together, these works cover important parts of the pipeline, but, to our knowledge, none evaluates the complete process from a user request through live web search to the delivery of a usable 3D world.

The resulting capability is relevant to applications such as virtual tourism and remote presence, site-specific embodied-agent training, cultural-heritage visualization, and AR/VR content creation, where a system must connect web evidence to a spatially coherent 3D or dynamic world. Recent work provides complementary building blocks: system-level intelligence flow across heterogeneous resources~\cite{an2025aiflowperspectivesscenarios,shao2025aiflownetworkedge}, single-image 3D lifting and metric scene recovery~\cite{chen2024cascadezero123imagehighlyconsistent,chen2024liftimage3dliftingsingleimage,wen2025metricsolverslidinganchoredmetric}, video generation and control~\cite{chen2025teleworlddynamicmultimodalsynthesis,xiang2025macrofrommicroplanninghighqualityparallelized,xiang2026pathwisetesttimecorrectionautoregressive,xiang2025makeefficientdynamicsparse,huang2025zero,huang2024domainfusion,Zhang_2026_CVPR}, and 4D/world generation~\cite{chen2026full4dgeneratingfullscope4d,zhang2026physomniphysicsgroundedmultiobjectscene,zhang2026tourphysicsbringingphysicsworld,wang2026directingworldfastautoregressive,xiang2026videoweaveunlockinggeometricconsistency,huang2026cineweavertrainingfreereferencecontrollablemultishot}. These capabilities motivate evaluating whether there exists a complete pipeline can turn a user request into a delivered 3D world.

\begin{figure*}[!t]
  \centering
  \includegraphics[width=\textwidth]{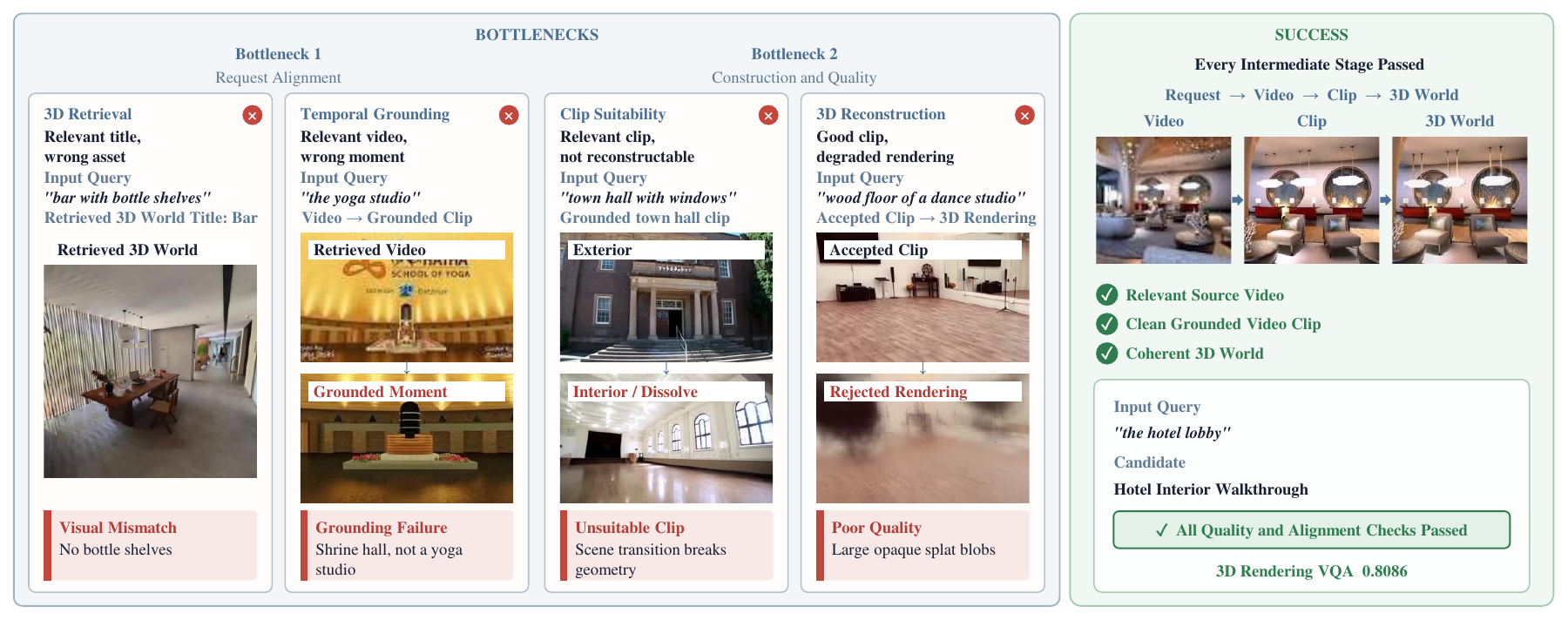}
  \caption{\textbf{Intermediate-stage bottlenecks and a successful \evaluation\ path.} The left side shows how incorrect 3D retrieval, temporal grounding, clip suitability, or reconstruction quality can prevent usable 3D world delivery. The right side shows a successful path from a relevant video through a clean grounded clip to a coherent 3D world that satisfies the request.}
  \label{fig:artifact}
\end{figure*}

This process motivates three coupled questions. First, how can an agentic system reliably transform relevant and accessible web-sourced visual content into a usable 3D world? Second, how should success be evaluated at different stages, distinguishing relevant retrieval from delivery of a request-aligned and perceptually acceptable world? Third, how can we resolve bottlenecks happened in the pipeline, such as visual mismatch, grounding failure, unsuitable clips for reconstruction, and the reconstructed 3D world of poor quality, as illustrated in Figure~\ref{fig:artifact}?



To answer the first two questions, we introduce \evaluation, an evaluation task that defines an end-to-end pipeline from a user request through web-sourced visual content retrieval to the delivery of a usable 3D world. Then, we design two metrics: Observed Retrieval Rate (ORR) measures whether a relevant 3D world resource or corresponding video resource is retrieved during the process, while World Delivery Rate (WDR) measures end-to-end 3D world delivery fulfillment. Together, ORR and WDR distinguish observed relevant retrieval opportunity from successful world delivery.

To address the third question, we propose \method, a reuse-then-reconstruction harness system with structured recovery control. It first seeks an acceptable reusable 3D world. If no such 3D world is found, it retrieves a source video, grounds a request-relevant temporal interval, and reconstructs a 3D world. When reconstruction fails, the harness can call subagents to revise the temporal grounding, substitute the source video, or reformulate the search and grounding queries. This design helps to solve the bottlenecks, that may be encountered in the whole pipeline. With these modules, \method\ combines efficient world reuse with controlled reconstruction fallback and recovery within a single execution.


We evaluate the mainstream models under the same \method\ setting on 300 scored requests spanning 20 fine-grained semantic domains, indoor and outdoor environments, and 5 spatial scales. A separate pool of 8,900 requests generated by the same data engine supports teacher-driven trajectory generation and supervised fine-tuning of the recovery agents. The results show that the agent model affect the overall performance of \evaluation\ task, that observing relevant visual content does not ensure successful world delivery, and that jointly training the recovery agents improves delivery and action efficiency.

The contributions of this paper are summarized as follows:
\begin{itemize}
    \item We introduce the \evaluation\ task, which defines an end-to-end pipeline from a user request through web-sourced visual content acquisition to usable 3D world delivery.
    \item We design two metrics for \evaluation\ task, ORR and WDR, to distinguish observed retrieval opportunity from successful world delivery.
    \item We present \method, a reuse-then-reconstruction harness system with structured recovery control that enables efficient completion of the \evaluation\ task while robustly handling failures encountered during execution.
\end{itemize}

\section{Related Work}
\label{sec:related}

The works most relevant to \evaluation\ fall into three lines: web content retrieval and acquisition, 3D reconstruction and generation, and interactive world model evaluation. Figure~\ref{fig:evaluation_boundary} organizes these lines by their starting inputs, used actions, and terminal outputs. Web-agent and multimodal search evaluation reaches web states, answers, reports, or grounded content. 3D reconstruction and generation evaluation begins from supplied visual content, conditioning signals, or candidate outputs, and interactive world model evaluation begins with an already instantiated world. These lines provide complementary capabilities, while \evaluation\ connects a user request, web-sourced visual content acquisition, and usable 3D world delivery within one request-level execution pipeline. This connection is why direct 3D / world generation or evaluation within an already instantiated world does not replace \evaluation\ studied here.
Table~\ref{tab:benchmark_scope} provides a comparison of the evaluation families. It distinguishes whether each evaluation family covers web-sourced visual content, delivered 3D worlds, or interactive world rollouts, making the boundary between web-sourced visual content acquisition, world delivery, and behavior within a world explicit.

\begin{table*}[!t]
\centering
\scriptsize
\caption{\textbf{Evaluation boundary for representative evaluation families.} We compare starting points and whether a benchmark evaluates web-sourced visual content, delivered 3D worlds, or interactive world rollouts. For \evaluation, the evaluated dimensions follow the request-level execution pipeline from a user request, through web-sourced visual content, to usable 3D world delivery. Interactive-world rollout provides a complementary boundary by evaluating behavior after a world has already been instantiated. \cmark, \pmark, and \xmark indicate core, partial, and no coverage, respectively.}
\label{tab:benchmark_scope}
\setlength{\tabcolsep}{1.8pt}
\renewcommand{\arraystretch}{1.0}
\resizebox{\textwidth}{!}{%
\begin{tabular}{@{}>{\raggedright\arraybackslash}m{2.55in} >{\raggedright\arraybackslash}m{2.10in} >{\centering\arraybackslash}m{0.71in} >{\centering\arraybackslash}m{0.71in} >{\centering\arraybackslash}m{0.71in}@{}}
\toprule
\textbf{Evaluation family} & \textbf{Starting Point} &
\shortstack{\textbf{Live-Web}\\\textbf{Content}} &
\shortstack{\textbf{Delivered}\\\textbf{World eval.}} &
\shortstack{\textbf{Interactive}\\\textbf{World Rollout}} \\
\midrule
Web Agent eval.
\cite{zhou2024webarena,koh2024visualwebarena,wei2025browsecomp,yoran2024assistantbench,gou2025mind2web2,wang2026liveresearchbench,gao2026drarena}
& Request + Live Web & \cmark & \xmark & \xmark \\
Multimodal Search eval.
\cite{jiang2024mmsearch,tao2025mmsearchplus,du2026mmsearchplus,liang2025videobrowser,liu2026videodr,yu2026rvms,zhang2025deepvideodiscovery,chen2026browsecompplus,xie2026dr3eval}
& Request + Multimodal Information + Live Web & \cmark & \xmark & \xmark \\
3D Reconstruction eval.
\cite{jensen2014largescale,schops2017multiview,knapitsch2017tanks,liang2024nvsquality,martin2025gsqa}
& Multi-View Images / Video & \xmark & \cmark & \xmark \\
3D / World Gen eval.
\cite{he2023t3bench,duggal2025eval3d,maiti2025gen3deval,wu2024gpteval3d,tam2026sceneeval,fan2026geot2v,li2025worldmodelbench,duan2025worldscore,qin2025worldsimbench,lu2025fourworldbench}
& Text / Condition & \xmark & \cmark & \pmark \\
Interactive-World eval.
\cite{wbench2026,worldroambench2026,worldexam2026,worldolympiad2026,playworld2026,harnessevalw2026}
& World State + Action / Camera & \xmark & \pmark & \cmark \\
\rowcolor{blue!8}
\textbf{\evaluation\ (Ours)}
& \textbf{Request + Live Web}
& \textbf{\cmark} & \textbf{\cmark}
& \textbf{\xmark} \\
\bottomrule
\end{tabular}
}
\end{table*}

\subsection{Web Content Retrieval and Acquisition Evaluation}

Web-agent and deep-research benchmarks make the user request the unit of evaluation while measuring web interaction, information seeking, and answer or report production. WebArena~\cite{zhou2024webarena} and VisualWebArena~\cite{koh2024visualwebarena} evaluate agents on realistic interactive web tasks. BrowseComp~\cite{wei2025browsecomp}, AssistantBench~\cite{yoran2024assistantbench}, and Mind2Web~2~\cite{gou2025mind2web2} evaluate browsing and citation-backed information seeking.
LiveResearchBench~\cite{wang2026liveresearchbench} and DR-Arena~\cite{gao2026drarena} evaluate user-centric deep-research reports over fresh or current sources. Their terminal outcomes are web states, verified answers, or reports.

Multimodal search benchmarks extend web retrieval to multimodal content information retrieval across images, text, and video. MMSearch~\cite{jiang2024mmsearch}, MMSearch-Plus~\cite{tao2025mmsearchplus}, and the long-horizon multimodal search benchmark~\cite{du2026mmsearchplus} evaluate provenance-aware browsing and cross-modal reasoning. Video-Browser~\cite{liang2025videobrowser}, Video Deep Research~\cite{liu2026videodr}, RVMS~\cite{yu2026rvms}, and Deep Video Discovery~\cite{zhang2025deepvideodiscovery} evaluate open-web video browsing, video reasoning, and temporal or moment localization, while BrowseComp-Plus~\cite{chen2026browsecompplus} and $DR^3$-Eval~\cite{xie2026dr3eval} provide controlled web settings. These benchmarks characterize web-sourced visual content acquisition through endpoints such as grounded answers, videos, and video moments. \evaluation\ carries the request beyond these endpoints by assessing whether the acquired visual content becomes a usable 3D world that fulfills the request.

\subsection{3D Reconstruction and Generation Evaluation}

3D reconstruction and generation evaluation typically begins once visual content, a conditioning signal, or a candidate output is available. Tanks and Temples~\cite{knapitsch2017tanks} evaluates large-scale reconstruction from supplied multi-view imagery. Output-oriented benchmarks assess generated 3D content: $T^3$Bench~\cite{he2023t3bench}, Eval3D~\cite{duggal2025eval3d}, Gen3DEval, GPT-4V(ision), and SceneEval cover text alignment, perceptual quality, geometric consistency, semantic coherence, and spatial plausibility \cite{he2023t3bench,duggal2025eval3d,maiti2025gen3deval,wu2024gpteval3d,tam2026sceneeval}. GeoT2V-Bench and GS-QA focus on reconstruction-based video consistency and Gaussian-splat view quality \cite{fan2026geot2v,martin2025gsqa}, while WorldModelBench, WorldScore, WorldSimBench, and 4DWorldBench extend evaluation to controllability, dynamics, simulation behavior, and 3D/4D consistency \cite{li2025worldmodelbench,duan2025worldscore,qin2025worldsimbench,lu2025fourworldbench}.

These benchmarks evaluate the reconstruction or generated world after the required visual content, conditioning signal, or candidate output has been supplied. \evaluation\ defines a completely different task: it starts with a user request, performs web-sourced visual content acquisition, and evaluates whether the complete execution produces a usable 3D world. Direct 3D / world generation therefore assesses a different starting condition and outcome rather than replacing the web-sourced visual content acquisition and delivery boundary measured by \evaluation.

\subsection{Interactive World Model Evaluation}

Interactive world model benchmarks evaluate behavior once a world has already been instantiated. WBench and WorldRoamBench study multi-turn interaction, consistency, memory, and long-horizon stability from supplied starting conditions \cite{wbench2026,worldroambench2026}. WorldExam and WorldOlympiad organize evaluation around appearance, reactivity, physical and geometric consistency, and interaction fidelity \cite{worldexam2026,worldolympiad2026}. PlayWorld uses a high-level objective and an agent player to drive a closed-loop rollout \cite{playworld2026}, while HarnessEval-W uses an agentified evaluator to diagnose existing visual-world rollouts \cite{harnessevalw2026}. Despite these different uses of agents, their starting point is a world state, interaction interface, or generated rollout, and their endpoint is an interactive rollout. This line provides a complementary boundary around behavior within an already instantiated world, while \evaluation\ measures the end-to-end execution pipeline from web-sourced visual content acquisition to 3D world delivery.

To sum up, prior benchmarks assess web-sourced visual content acquisition, 3D reconstruction or generation from supplied inputs, or behavior within an already instantiated world. \evaluation\ defines an end-to-end pipeline: it starts from a user request, acquires web-sourced visual content, and assesses the resulting 3D world. By reporting ORR and WDR, it separates observed retrieval opportunity from successful world delivery.

\section{\evaluation}
\label{sec:evaluation}

\begin{figure*}[t]
  \centering
  \includegraphics[width=\textwidth,pagebox=cropbox]{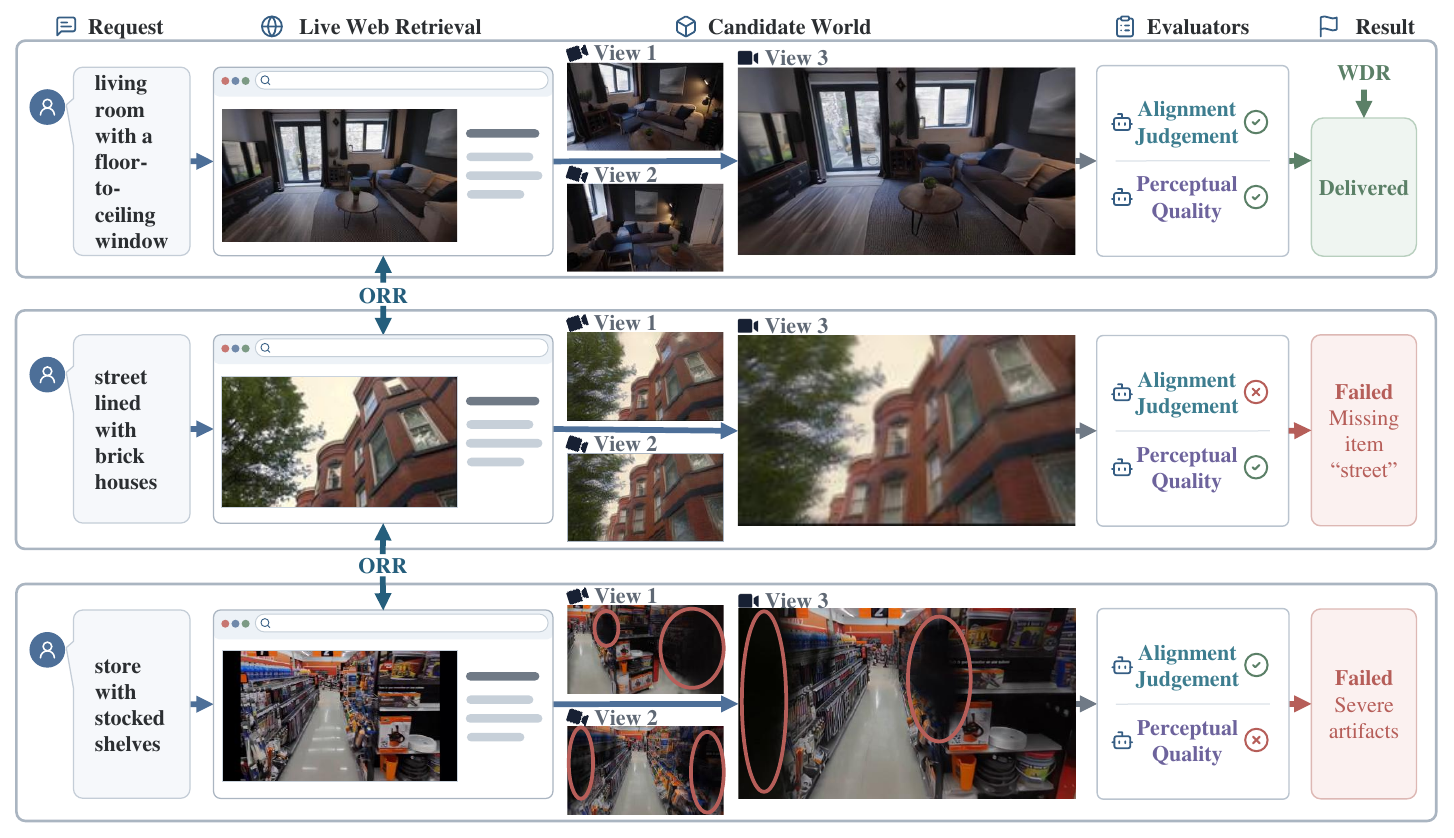}
  \caption{\textbf{Overview of the \evaluation\ task workflow and typical examples.} \evaluation\ defines an end-to-end execution pipeline from a user request through web-sourced visual content acquisition and candidate 3D world delivery to automatic judgments of request alignment and perceptual quality. Observed Retrieval Rate (ORR) records whether relevant candidate web-sourced visual content is observed during acquisition, whereas World Delivery Rate (WDR) counts a request as delivered only when the resulting world satisfies both criteria.}
  \label{fig:search_to_scene_benchmark_overview}
\end{figure*}

\evaluation\ propose an end-to-end execution pipeline from a user request to a usable 3D world and an evaluation benchmark upon it. It separates visual content retrieval opportunity from final delivery fulfillment: Observed Retrieval Rate (ORR) records whether relevant candidate web-sourced visual content is observed during acquisition, whereas World Delivery Rate (WDR) counts a request as delivered only when the resulting 3D world satisfies both request alignment and perceptual quality. Figure~\ref{fig:search_to_scene_benchmark_overview} provides an overview of the \evaluation\ task workflow and illustrates representative examples.

\subsection{Task Definition}
\label{sec:task}

\evaluation\ asks whether an agentic system can turn request-aligned visual content discovered on the live web into a deliverable 3D world for a user request. The task begins with the request and ends only when the system produces a world that matches the requested content and spatial extent and meets the perceptual-quality criterion.

Each execution searches the live web for visual content needed for world delivery rather than querying a fixed knowledge base. Delivery from web-sourced visual content matters when a user request refers to specific real-world content that must be reflected in the delivered 3D world. Unlike 3D / world generation from a supplied condition, the evaluated system must discover corresponding visual content on the live web and carry it through to 3D world delivery. This correspondence supports downstream settings tied to specific real environments.

Let $\mathcal{F}$ denote the system, $q$ denote the request, and $\pi$ denote the complete system configuration. Let $\mathcal{W}$ denote the space of usable 3D worlds, and let $\mathcal{W}_\varnothing\notin\mathcal{W}$ denotes a failure of the task. The system output is
\begin{equation}
  w_{\mathrm{out}}=\mathcal{F}_{\pi}(q),
  \qquad
  w_{\mathrm{out}}\in\mathcal{W}\cup\{\mathcal{W}_\varnothing\},
  \label{eq:task}
\end{equation}
where $w_{\mathrm{out}}$ is either a candidate world in $\mathcal{W}$ or $\mathcal{W}_\varnothing$. The system separately records an execution trace, which supports metric computation and trajectory replay but does not enter the delivery label.

\subsection{Execution Pipeline}
\label{sec:request_to_world_path}

Figure~\ref{fig:search_to_scene_benchmark_overview} operationalizes the task as an end-to-end execution pipeline from user request to 3D world. Given a user request, an agentic system performs web-sourced visual content acquisition and produces a candidate 3D world. The candidate may be obtained as a reusable 3D world discovered during web-sourced visual content acquisition or reconstructed from a corresponding video resource after temporal grounding. The candidate then enters the automatic world delivery assessment, and the execution is delivered only when the world satisfies both request alignment and perceptual quality.

Executing these stages end-to-end requires more than observing relevant visual content. As Figure~\ref{fig:artifact} shows, intermediate-stage bottlenecks arising from incorrect 3D retrieval, temporal grounding, clip suitability, or reconstruction quality can interrupt the execution before world delivery. Section~\ref{sec:method} introduces \method\ to address these bottlenecks through a reuse-then-reconstruction harness system with structured recovery control.

\subsection{Automatic World Delivery Assessment}
\label{sec:acceptance_delivery}

The automatic world delivery assessment applies two Vision-Language Model (VLM)-based evaluators to multiview renderings of each candidate world, following previous output-oriented 3D evaluation benchmarks that assess textual alignment, perceptual quality, and rendered-view quality~\cite{he2023t3bench,duggal2025eval3d,maiti2025gen3deval,wu2024gpteval3d,martin2025gsqa}. The alignment evaluator assigns $P_{\mathrm{align}}(w,q)\in\{0,1\}$ to assess the relevance between the request $q$ and the renderings of world $w$. The quality evaluator first gives $Q_{\mathrm{qual}}(w)\in\mathbb{R}$ from the rendered visual content alone to assess the visual quality of world $w$ and then assigns
\begin{equation}
  P_{\mathrm{qual}}(w)
  =\mathbf{1}\{Q_{\mathrm{qual}}(w)\geq\tau_{\mathrm{qual}}\}.
\end{equation}
Together, the two labels define the world delivery indicator. Specifically, for a world $w$, the world delivery indicator can be denoted as
\begin{equation}
Y_{\mathrm{delivery}}(q)
=
\begin{cases}
P_{\mathrm{align}}(w,q)\,
P_{\mathrm{qual}}(w),
& w\in\mathcal{W},\\
0, & w=\mathcal{W}_\varnothing,
\end{cases}
\label{eq:delivery}
\end{equation}
The system output $\mathcal{F}_{\pi}(q)$ is therefore
\begin{equation}
  \mathcal{F}_{\pi}(q)=w\cdot Y_{\mathrm{delivery}}(q).
  \label{eq:delivered_output}
\end{equation}
Thus, a run receives a positive delivery label only when there exists a final world output aligning with the request and surpassing the perceptual-quality threshold.
Implementation details of the automatic world delivery assessment are in the \textbf{Harness Configurations} part of Section~\ref{sec:setup}.

\subsection{Evaluation Metrics}
\label{sec:outcomes}

Request-level task success is a standard outcome in agent evaluation because it measures whether the complete execution fulfills the request~\cite{zhou2024webarena,koh2024visualwebarena,ma2024agentboard,kapoor2024agents}. \evaluation\ adopts this end-to-end outcome through World Delivery Rate (WDR), the rate of positive delivery labels assigned by the automatic world delivery assessment. It additionally reports Observed Retrieval Rate (ORR), the proportion of runs that retrieve at least one relevant candidate item of visual content, which can be either a 3D world resource or a grounded video clip. Let $\mathcal{C}$ denote the set of candidate web-sourced visual content retrieved, and let $R(c,q)\in\{0,1\}$ denote the relevance assessment for candidate web-sourced visual content $c\in\mathcal{C}$ and request $q$, also given by a VLM evaluator. The retrieval indicator is
\begin{equation}
  O(q)=\mathbf{1}\!\left\{\exists c\in\mathcal{C}:
  R(c,q)=1\right\}.
  \label{eq:opportunity}
\end{equation}

Let $\mathcal{Q}$ be the evaluation set and $N=|\mathcal{Q}|$. For configuration $\pi$, the metrics are
\begin{align}
  \mathrm{WDR}(\pi)
  &=\frac{1}{N}\sum_{q\in\mathcal{Q}}Y_{\mathrm{delivery}}(q),
  \label{eq:WDR}\\
  \mathrm{ORR}(\pi)
  &=\frac{1}{N}\sum_{q\in\mathcal{Q}}O(q).
  \label{eq:orr}
\end{align}
WDR measures successful world delivery, whereas ORR measures observed retrieval opportunity. This task-specific separation distinguishes observing relevant visual content from fulfilling the world request.

\section{\method: Reuse-Then-Reconstruction}
\label{sec:method}

\begin{figure}[t]
  \centering
  \includegraphics[width=0.7\linewidth]{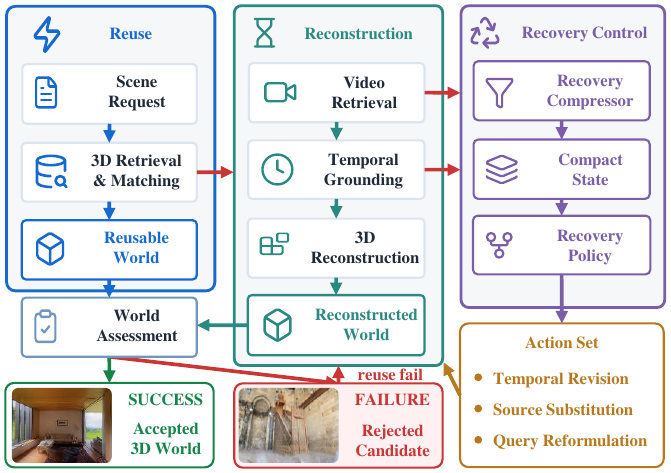}
  \caption{\textbf{\method\ harness system.} The system first seeks a reusable 3D world and falls back to the reconstruction branch when no reusable world is accepted. The reconstruction branch executes video retrieval, temporal grounding, and 3D reconstruction to fulfill the final 3D world delivery. When the reconstruction branch encounters a failure, the recovery compressor agent updates a compact recovery state, and the recovery policy agent selects a temporal-revision, source-substitution, or query-reformulation action. The action set enumerates these three recovery-action types. At step $t$, the harness restricts them to the feasible set $\mathcal{A}(x_t)$, from which the policy selects $a_t\in\mathcal{A}(x_t)$. \textcolor[HTML]{C93636}{Red arrows} denote failure-triggered transitions.}
  \label{fig:worldsearcher_overview}
\end{figure}

The \evaluation\ task defined in Section~\ref{sec:request_to_world_path} exposes intermediate-stage bottlenecks in 3D retrieval, temporal grounding, clip suitability, and reconstruction quality, as illustrated in Figure~\ref{fig:artifact}. To address these bottlenecks, \method\ realizes the \evaluation\ task as a reuse-then-reconstruction harness system with structured recovery control. Given a user request, the harness first seeks a reusable 3D world and invokes video-based reconstruction when no reusable world is accepted. Within the reconstruction branch, the recovery compressor agent updates a compressed recovery state after an observed runtime failure, and the recovery policy agent selects a state-admissible recovery action.
Figure~\ref{fig:worldsearcher_overview} summarizes the \method\ execution flow, including the reuse branch, reconstruction branch, and recovery control loop.

\subsection{Reuse-Then-Reconstruction}
\label{sec:overview}

Inspired by the reuse-before-recomputation principle underlying cache design in computer architecture, \method\ organizes world acquisition as a reuse-then-reconstruction system. Rather than reconstructing every 3D world from raw web-sourced visual content, which is expensive, it first seeks an existing 3D world and accepts it as reusable only when it passes the automatic world delivery assessment. It invokes video-based reconstruction when no candidate reusable 3D world is accepted, which is the visual mismatch bottleneck shown in Figure~\ref{fig:artifact}.

In the reconstruction branch, \method\ retrieves web video, locates a request-relevant temporal interval, and reconstructs a 3D world. Thus, the reuse branch provides the first world-acquisition route, while the reconstruction branch supplies the fallback for world delivery.

\subsection{Structured Recovery Control}
\label{sec:recovery_control}

As illustrated in Figure~\ref{fig:artifact}, bottlenecks like grounding failure, unsuitable clips for reconstruction, and the reconstructed 3D world of poor quality may happen during runtime of the reconstruction branch. Therefore, we design a structured recovery control to convert observed runtime failures into state-admissible recovery decisions. It comprises a recovery compressor agent, a compressed recovery state, a harness-enforced feasible action set, and a recovery policy agent. The recovery compressor agent updates $m_t$ from the observed failure and prior guidance, while the recovery policy agent selects the next action $a_t$ from the feasible action set $\mathcal{A}(x_t)$.

\noindent\textbf{Recovery Action Set.}
Within the reconstruction branch, the \method\ harness defines three actions that determine the scope of recovery after a failure. Each action specifies the portion of the current trajectory retained during recovery: the harness can try another clip within the same video, switch to another video under the same query, or reformulate the search and temporal grounding queries.

These actions operate at three levels of the visual-content-to-world process. The temporal revision action, $a_{\mathrm{temporal}}$, explores an alternative temporally grounded clip within the same video.  The source substitution action, $a_{\mathrm{source}}$, moves to another video supporting the same query. The query reformulation action, $a_{\mathrm{query}}$, revises the search and temporal grounding queries while preserving the requested world semantics. Together, the three actions provide temporal-, source-, and query-level recovery without imposing a fixed selection order.

Let $x_t$ denote the recovery state maintained by the harness at step $t$. From this state, the harness derives the feasible action set:
\begin{equation}
  \mathcal{A}(x_t)
  =
  \{a\in\{a_{\mathrm{temporal}},a_{\mathrm{source}},a_{\mathrm{query}}\}:
  \rho_a(x_t)=1\},
  \label{eq:admissible_atomic_actions}
\end{equation}
where $\rho_a(x_t)$ indicates whether action $a$ is compatible with state $x_t$. The feasible action set $\mathcal{A}(x_t)$ is an input constraint to the recovery policy rather than an output generated by it. This separation keeps recovery feasibility under harness control, while the recovery policy agent selects only from the admissible alternatives.

\noindent\textbf{Recovery Policy and Compressor.}
The \method\ harness coordinates the recovery compressor agent and recovery policy agent with complementary roles. Let $f_t$ denote the failure observed at recovery step $t$. The recovery compressor agent first updates the compressed recovery state, after which the recovery policy agent selects the next action from $\mathcal{A}(x_t)$:
\begin{align}
  m_t &= \Phi(q,x_t,m_{t-1},f_t),
  \qquad m_0=m_{\mathrm{init}},
  \label{eq:recovery_compressor}
  \\
  a_t &= \Pi(q,m_t;\mathcal{A}(x_t)).
  \label{eq:recovery_policy}
\end{align}
Here, $\Phi$ and $\Pi$ denote the recovery compressor agent and recovery policy agent, respectively. The variable $m_t$ denotes the compressed recovery state at step $t$, with $m_0=m_{\mathrm{init}}$ specifying the empty initial state, while $a_t\in\mathcal{A}(x_t)$ denotes the selected action. Accordingly, the recovery policy selects temporal revision, source substitution, or query reformulation only when that action is admissible in the current harness state.

The compressed state $m_t$ encodes prior guidance and the current failure in a fixed schema that summarizes the failure and records concepts to avoid, visual targets to preserve, and guidance for the next query. This representation provides the recovery policy agent with a compact recovery context without requiring replay of the full trajectory history, consistent with prior work on structured search states and compact iterative memory \cite{zhu2026retrac,yuan2026memsearcher}.

\section{Experiments}
\label{sec:experiments}

\subsection{Evaluation Settings}
\label{sec:setup}

\textbf{Evaluation Models.} For the full model comparison, we evaluate eight open-weight models: four models of different parameter sizes from the Qwen3.5 series \cite{qwen2026qwen35collection}, two models from the DeepSeek-V4 series \cite{deepseek2026v4}, GLM-5.2 \cite{zai2026glm52}, and Kimi K3 \cite{moonshot2026kimik3weights}. We also evaluate four closed-source models: GPT-5.6 Sol and GPT-5.6 Terra from the GPT-5.6 series \cite{openai2026gpt56}, Claude Fable 5 \cite{anthropic2026fable5}, and Claude Opus 5 \cite{anthropic2026opus5}. Within each full-model configuration, the query generation, query refinement, recovery policy, and recovery compressor agents all use the same evaluation model. For the recovery ablation, we use Qwen3.5-9B as the base model and train separate LoRA adapters for the recovery policy and recovery compressor agents with supervised fine-tuning (SFT).


\noindent\textbf{Evaluation Protocols.} For each evaluation case, every configuration receives the same user request and is evaluated on the same 300-request evaluation set. The remaining harness components, auxiliary models, and video-path action cap are fixed within each comparison. We use the same frozen alignment and quality evaluator instances described in Section~\ref{sec:acceptance_delivery} for all world-delivery labels; both evaluator instances use GPT-5.6 Sol. We report ORR, WDR, and mean actions used as primary outcomes. The sealed evaluation inventory, fixed harness interfaces, and comparison design are documented in the supplementary materials.

\noindent\textbf{Harness Configurations.} The \method\ harness coordinates four primary agents: query generation, query refinement, recovery policy, and recovery compression. It also uses the following auxiliary components: the Serper \cite{serper2026} search engine, the Qwen3-VL-Reranker \cite{li2026qwen3vlreranker}, the TimeLens temporal grounder \cite{zhang2026timelens}, the Q-ReAlign quality model \cite{qfuture2026qrealign}, and a GPT-5.6 Sol semantic matcher. The video-reconstruction branch uses an action cap of $B=36$. We report the Q-ReAlign score and the CLIP score of delivered 3D worlds in the supplementary materials to verify the effectiveness of the Q-ReAlign quality model and the GPT-5.6 Sol semantic matcher.

\subsection{Data Engine}
\label{sec:data_engine}

\begin{figure*}[!t]
  \centering
  \includegraphics[width=\textwidth,pagebox=cropbox]{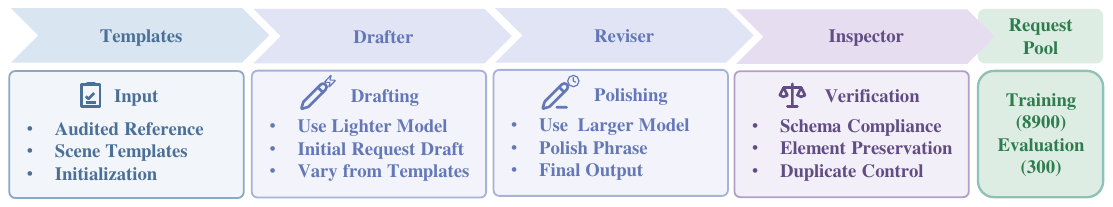}
  \caption{\textbf{Data construction pipeline.} Audited reference requests and paired scene templates provide initialization, which is then transformed into a controlled natural-language request dataset through drafting, polishing, and verification. Accepted requests form an 8,900-request training-source pool for teacher-driven SFT data generation and a 300-request evaluation set.}
  \label{fig:data_engine_construction}
\end{figure*}

To support controlled generation of SFT trajectories and end-to-end evaluation, we construct a data engine for request dataset generation. The engine starts from manually audited reference requests paired with scene templates and generates natural-language variants constrained to preserve the requested scene semantics. The construction pipeline separates drafting, polishing, and rubric-based verification into three functional roles:
\begin{itemize}
    \item \textbf{Drafter.} Apply a lighter model to quickly generate an initial request draft conditioned on the audited request and its associated scene template, while introducing controlled wording variation.
    \item \textbf{Reviser.} Apply a larger model for fine-grained linguistic refinement, rewriting each draft as a concise natural-language request while preserving the requested scene semantics.
    \item \textbf{Inspector.} Combine predefined rule-based checks with LLM verification to assess compliance with the request schema, preservation of required scene elements, and duplicate control.
\end{itemize}
In our implementation, GPT-5.6 Luna instantiates the Drafter, while GPT-5.6 Sol instantiates both the Reviser and the Inspector. Only candidates that pass the Judge's rubric enter the final request collection, as illustrated in Figure~\ref{fig:data_engine_construction}.

\begin{figure*}[!t]
  \centering
  \includegraphics[width=\textwidth,pagebox=cropbox]{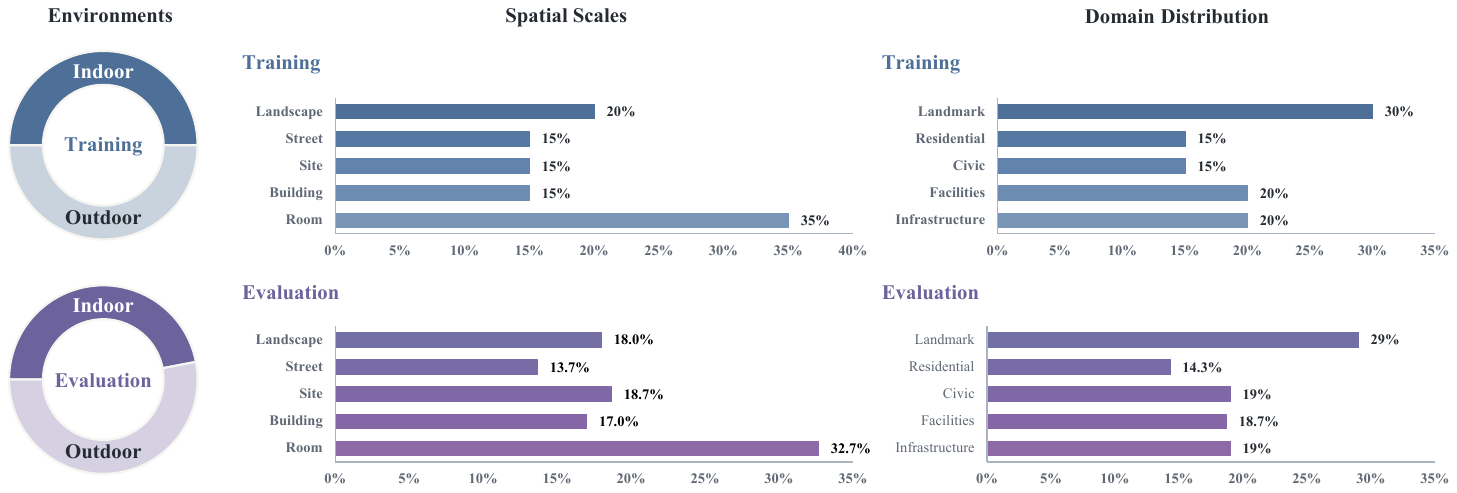}
  \caption{\textbf{Composition of the dataset.} Distribution of the 8,900-request training-source pool and the 300-request evaluation set across various environments, spatial scales, and 5 higher-level semantic domains that aggregate the 20 fine-grained domains.}
  \label{fig:data_engine_distribution}
\end{figure*}

\noindent\textbf{Data Composition.}
The resulting request collection contains an 8,900-request training-source pool for generating teacher trajectories used for SFT and a 300-request evaluation set. Both subsets cover 20 fine-grained semantic domains, grouped into 5 higher-level semantic domains. They also span indoor and outdoor environments and 5 spatial scales. Figure~\ref{fig:data_engine_distribution} shows the distribution of the two subsets across the semantic, environmental, and spatial dimensions.

The 8,900-request training-source pool contains 445 requests in each of 20 fine-grained domains, 4,450 indoor and 4,450 outdoor requests, and 3,115 room-level, 1,335 building-level, 1,335 site-level, 1,335 street-level, and 1,780 landscape-level requests. The 300-request evaluation set contains 13--23 requests per domain across all 20 domains, 141 indoor and 159 outdoor requests, and 98 room-level, 51 building-level, 56 site-level, 41 street-level, and 54 landscape-level requests.
More details of the dataset and the complete evaluation set are shown in the supplementary materials.

\subsection{Full Model Comparison}
\label{sec:model_comparison}

For the full model comparison experiment, we apply each candidate model to the four agents while keeping the harness configuration, delivery gates, and action budget fixed. This design measures how the model assignment affects the complete \evaluation\ task rather than an isolated agent stage. Results are shown in Table~\ref{tab:model_comparison}.

\begin{figure*}[htbp]
  \centering
  \includegraphics[width=0.99\textwidth,pagebox=cropbox]{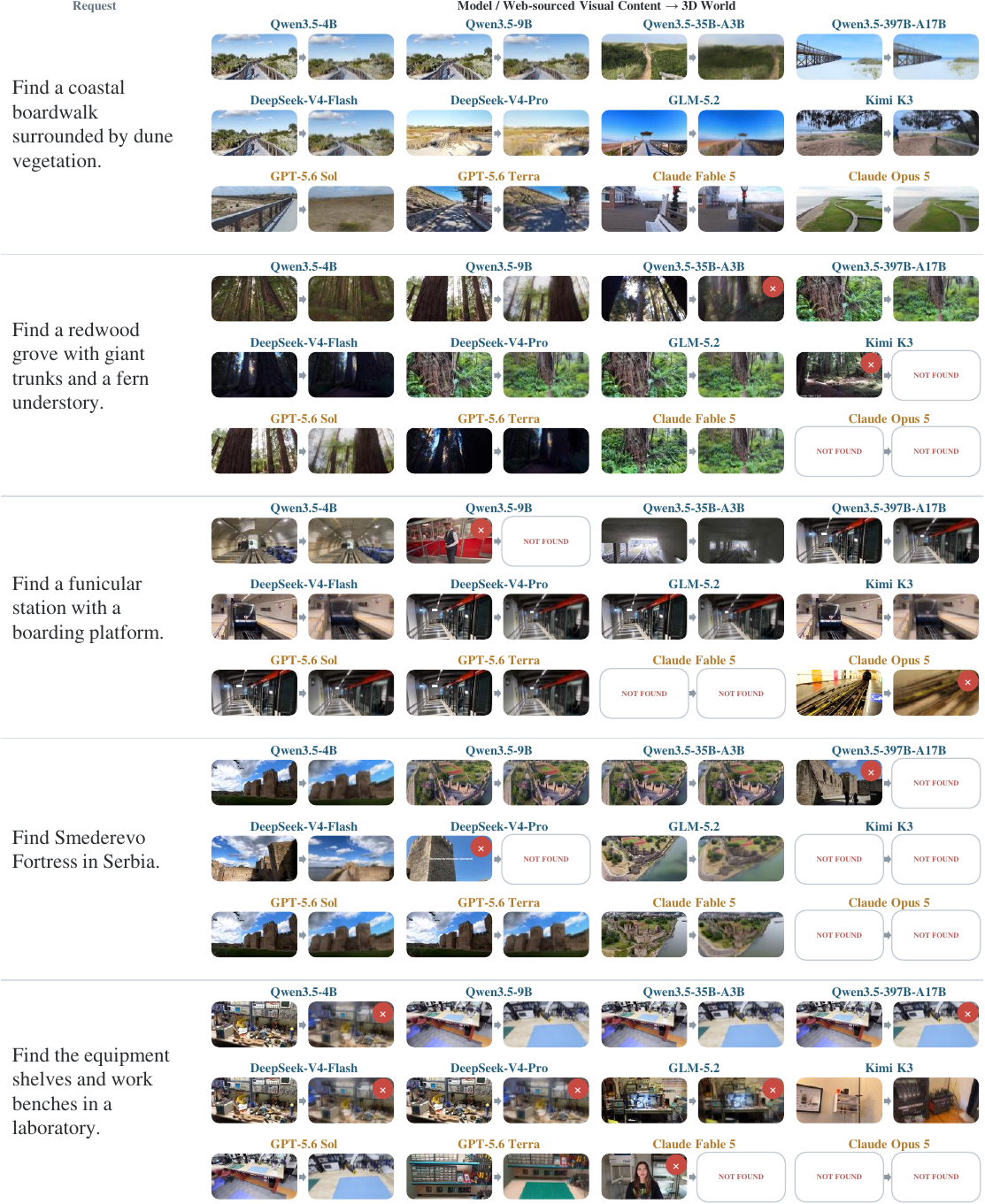}
  \caption{\textbf{Qualitative comparison across full-model configurations.} Each case block shows one request through the web-sourced visual content to 3D world for all evaluated models, tracing how model-specific trajectories propagate from request through web-sourced visual content retrieval to delivery. Successful result pairs retrieve video clips with the delivered 3D worlds, while stage-specific failures are shown in place. \textcolor[HTML]{C93636}{\texttt{NOT FOUND}} denotes an unavailable stage, and the \textcolor[HTML]{C93636}{red cross} denotes a stage result rejected by its corresponding quality gate. More visualizations of results generated by the reuse and reconstruction branch will be exhibited in the supplementary materials.}
  \label{fig:full_model_qualitative}
\end{figure*}

The results separate three useful operating profiles. DeepSeek-V4-Flash achieves the highest ORR (88.3\%) and WDR (78.7\%) across the evaluated configurations, while also using only 17.3 mean video-path actions. GLM-5.2 uses the fewest actions overall (13.9) and reaches 52.7\% WDR, showing a distinct efficiency-oriented profile. Within the closed-source group, GPT-5.6 Sol attains the highest WDR (53.0\%) with 23.2 mean actions, while GPT-5.6 Terra attains the highest ORR in that group (72.7\%). The variation between ORR and WDR across models shows that observing relevant visual content and completing an accepted world are complementary outcomes: model choice affects visual content acquisition from the live web, downstream delivery, and interaction cost together. The qualitative cases in Figure~\ref{fig:full_model_qualitative} make this distinction concrete by tracing where model-specific trajectories reach or leave the delivery path.

\begin{table}[!t]
\centering
\scriptsize
\caption{\textbf{Full model comparison.}
We apply each model to the four primary agents and keep the other configurations fixed. All models use the same 300-request evaluation set. We compare eight open-weight models: Qwen3.5 series, DeepSeek-V4 series, GLM-5.2, and Kimi K3. We also compare four closed-source models: GPT-5.6 Sol, GPT-5.6 Terra, Claude Fable 5, and Claude Opus 5. ORR (Observed Retrieval Rate) is the percentage of runs that retrieve at least one relevant direct 3D scene or corresponding video clip candidate.
WDR (World Delivery Rate) is the percentage of runs that satisfy request alignment, spatial extent, and perceptual quality at the delivery endpoint.
Mean Actions is the mean number of video-path actions used per run.
ORR and WDR are reported as percentages; Mean Actions is reported as actions per run.
\textbf{Bold}: best observed value. \underline{Underline}: second-best observed value.}
\label{tab:model_comparison}
\setlength{\tabcolsep}{1.5pt}
\begin{tabular*}{\columnwidth}{@{\extracolsep{\fill}}lccc@{}}
\toprule
Model & ORR (\%) $\uparrow$ & WDR (\%) $\uparrow$ & Mean Actions $\downarrow$ \\
\midrule
\multicolumn{4}{l}{\textit{Open-weight models}} \\
Qwen3.5-4B & 42.7 & 38.0 & 24.9 \\
Qwen3.5-9B & 34.0 & 29.3 & 26.6 \\
Qwen3.5-35B-A3B & 49.3 & 39.7 & 24.0 \\
Qwen3.5-397B-A17B & 68.7 & 44.0 & 25.4 \\
DeepSeek-V4-Flash & \textbf{88.3} & \textbf{78.7} & \underline{17.3} \\
DeepSeek-V4-Pro & 63.7 & 48.3 & 22.0 \\
GLM-5.2 & 66.0 & 52.7 & \textbf{13.9} \\
Kimi K3 & 54.0 & 52.7 &19.9 \\
\addlinespace
\multicolumn{4}{l}{\textit{Closed-source models}} \\
GPT-5.6 Sol & 68.0 & \underline{53.0} & 23.2 \\
GPT-5.6 Terra & \underline{72.7} & 51.0 & 24.1 \\
Claude Fable 5 & 41.0 & 29.3 & 29.1 \\
Claude Opus 5 & 23.7 & 16.0 & 32.8 \\
\bottomrule
\end{tabular*}
\end{table}

\subsection{Recovery Ablation}
\label{sec:reuse_diagnostics}

\noindent\textbf{Teacher-Trajectory Generation.}
We use the data engine in Section~\ref{sec:data_engine} to construct the 8,900-request training-source pool, which matches the evaluation set in semantic, environmental, and spatial-scale coverage. For each request in this pool, we execute the \method\ harness with GPT-5.6 Sol as the teacher model and record the resulting runtime trajectory.

\noindent\textbf{SFT Training.}
Using these teacher-generated trajectories, we train two independent LoRA adapters initialized from Qwen3.5-9B with MS-Swift \cite{zhao2024swiftascalablelightweightinfrastructure}: one for the recovery policy agent and one for the recovery compressor agent. The training targets follow the output schemas used at inference; policy targets are additionally restricted to the state-conditioned admissible action set. For policy SFT, we retain examples from trajectories whose final output receives a positive world-delivery label. For compressor SFT, we retain state-update examples whose outputs conform to the fixed recovery-state schema and faithfully encode the observed runtime failure and relevant recovery context. This role-specific filtering aligns the supervision with each agent's inference-time contract.

For the recovery ablation, we evaluate three SFT variants that isolate the individual and joint contributions of the two adapters.
\begin{itemize}
    \item \textbf{Policy-SFT Only.} The recovery policy agent uses the Policy-SFT adapter, while the recovery compressor and all other primary agents use GPT-5.6 Sol.
    \item \textbf{Compressor-SFT Only.} The recovery compressor agent uses the Compressor-SFT adapter, while the recovery policy and all other primary agents use GPT-5.6 Sol.
    \item \textbf{Both SFT.} The recovery policy agent and recovery compressor agent use the Policy-SFT and Compressor-SFT adapters, respectively. All other primary agents use GPT-5.6 Sol.
\end{itemize}

We additionally report two reference configurations. One uses GPT-5.6 Sol for both recovery agents, whereas the other uses the Qwen3.5-9B base model for both. Results are reported in Table~\ref{tab:recovery_results}.

\begin{table}[!t]
\centering
\scriptsize
\caption{\textbf{Recovery ablation.}
All variants use the same configurations. Only the recovery-policy and recovery-compressor model assignments vary.
Policy-SFT and Compressor-SFT denote supervised fine-tuning for the corresponding recovery agent, while Both SFT uses both adapters.}
\label{tab:recovery_results}
\setlength{\tabcolsep}{1.5pt}
\begin{tabular*}{\columnwidth}{@{\extracolsep{\fill}}lccc@{}}
\toprule
Model & ORR (\%) $\uparrow$ & WDR (\%) $\uparrow$ & Mean Actions $\downarrow$ \\
\midrule
Teacher Model & \textbf{68.0} & \textbf{53.0} & \textbf{23.2} \\
Base Model & 49.0 & 40.3 & 26.6 \\
Policy-SFT Only & 57.0 & 44.3 & 25.8 \\
Compressor-SFT Only & 52.3 & 42.0 & 26.2 \\
Both SFT & \underline{64.0} & \underline{49.7} & \underline{24.5} \\
\bottomrule
\end{tabular*}
\end{table}

This recovery ablation evaluates the recovery control design of \method\ by isolating its two structured recovery subagents and quantifying the benefits of role-specific SFT. Relative to the base model, the Policy-SFT Only variant raises ORR from 49.0\% to 57.0\% and WDR from 40.3\% to 44.3\%, while reducing the mean actions used from 26.6 to 25.8, consistent with more effective selection of admissible recovery actions. The Compressor-SFT Only variant likewise raises ORR from 49.0\% to 52.3\% and WDR from 40.3\% to 42.0\%, consistent with improved compliance with the recovery-state schema and more faithful preservation of decision-relevant failure information. Using both adapters yields the strongest SFT configuration, reaching 64.0\% ORR, 49.7\% WDR, and a mean action count of 24.5. These joint gains support the intended complementary control structure: compressor SFT improves the fidelity of the structured recovery state, while policy SFT improves the selection of an admissible recovery action conditioned on that state.

\section{Conclusion}

We introduce \evaluation, an evaluation task that defines an end-to-end pipeline that can turn a user request into a usable 3D world through web-sourced visual content. By using two separate metrics, ORR and SDR, \evaluation\ makes visible the gap between retrieving relevant visual content from the live web and completing accepted world delivery. We also present \method, a reuse-then-reconstruction harness system that combines direct 3D reuse, video-based reconstruction, and structured recovery. Across 300 requests used in the evaluation experiment, our results show that the agent model affects retrieval, delivery, and action cost. Jointly adapting the recovery policy and compressor yields the strongest SFT recovery variant, demonstrating their complementary roles. Together, \evaluation\ and \method\ establish a measurable and effective path from user requests to 3D world delivery.

\clearpage
\appendix
\graphicspath{{assets/supplementary/}{assets/figures/}}
\setlength{\parindent}{1em}
\setlength{\parskip}{0pt}
\setlength{\emergencystretch}{1.5em}
\setlength{\tabcolsep}{1.5pt}
\renewcommand{\arraystretch}{1.05}
\lstset{basicstyle=\ttfamily\scriptsize,breaklines=true,breakatwhitespace=false,columns=fullflexible,frame=single,framerule=0.3pt,xleftmargin=3pt,xrightmargin=3pt,aboveskip=4pt,belowskip=6pt,keepspaces=true,showstringspaces=false}
\tcbset{
  promptbase/.style={
    enhanced,
    breakable=true,
    listing only,
    colback=black!4,
    colframe=black!55,
    colbacktitle=black!62,
    coltitle=white,
    fonttitle=\bfseries\small,
    boxrule=0.5pt,
    arc=2pt,
    left=4pt,
    right=4pt,
    top=3pt,
    bottom=3pt,
    before skip=5pt,
    after skip=7pt,
    listing options={
      basicstyle=\ttfamily\footnotesize,
      breaklines=true,
      breakatwhitespace=false,
      columns=fullflexible,
      keepspaces=true,
      showstringspaces=false,
      xleftmargin=0pt,
      xrightmargin=0pt
    }
  }
}
\newtcblisting{systemprompt}{promptbase,title={System prompt}}
\newtcblisting{userprompt}{promptbase,title={User prompt}}
\newtcblisting{promptpart}[1]{promptbase,title={#1}}
\begin{center}
  {\Large\bfseries Supplementary Materials\par}
\end{center}

\section{Visualization and metrics of Delivered 3D World}

\subsection{Reuse 3D World Delivery}

\begin{figure}[htbp]\centering
\includegraphics[width=\linewidth,height=0.92\textheight,keepaspectratio]{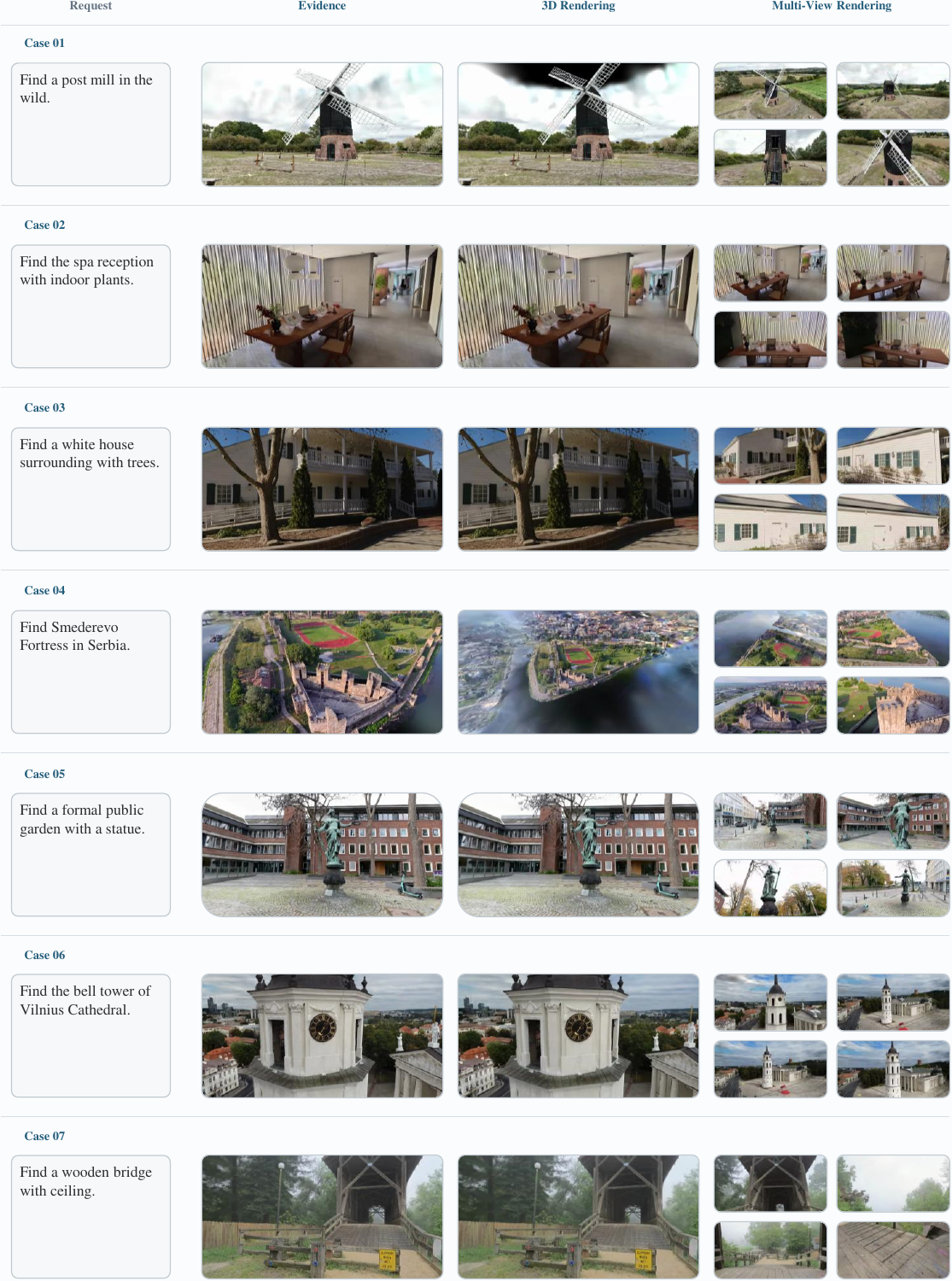}
\caption{Visualization of Reuse Result.}
\label{fig:reuse}
\end{figure}

Figure \ref{fig:reuse} shows several representative deliveries from the reuse branch. Each row pairs the user request with the retrieved 3D resource and several multi-view renderings.

\subsection{Reconstruction 3D World Delivery}
\begin{figure}[htbp]\centering
\includegraphics[width=\linewidth,height=0.92\textheight,keepaspectratio]{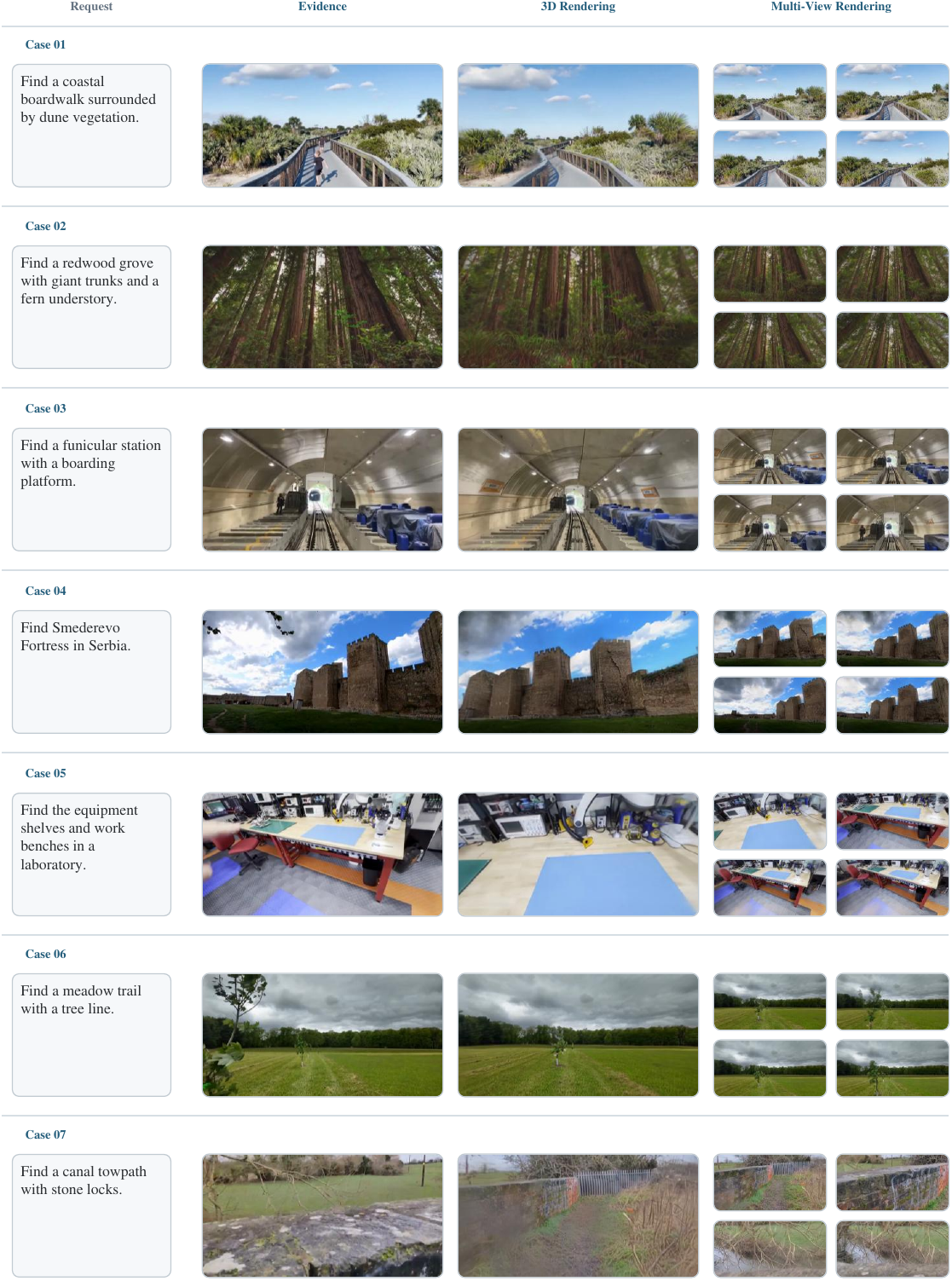}
\caption{Visualization of Reconstruction Result.}
\label{fig:recon}
\end{figure}

Figure \ref{fig:recon} shows several representative deliveries from the reconstruction branch. Each row pairs the user request with the retrieved video resource and several multi-view renderings.

\subsection{Metrics of Delivered 3D World}

We report some additional metrics to evaluate the quality and alignment of the delivered 3D worlds, including PSNR \cite{hore2010image}, LPIPS \cite{zhang2018unreasonable}, SSIM \cite{wang2004image}, CLIP Score \cite{hessel2021clipscore}, and Q-ReAlign \cite{qfuture2026qrealign}.
Table~\ref{tab:final_render_quality_summary} reports metrics of the delivered 3D world using GPT-5.6 Sol.
\begin{table}[htbp]\centering
\normalsize
\caption{Additional metrics of delivered 3D world.}
\label{tab:final_render_quality_summary}
\begin{tabularx}{\linewidth}{@{}>{\bfseries\raggedright\arraybackslash}p{0.13\linewidth}*{5}{>{\centering\arraybackslash}X}@{}}
\toprule
Metric & \textbf{PSNR} & \textbf{LPIPS} & \textbf{SSIM} & \textbf{CLIP Score} & \textbf{Q-ReAlign VQA} \\
\midrule
Case mean & 25.36 & 0.30 & 0.66 & 26.37 & 0.64 \\
\bottomrule
\end{tabularx}
\end{table}

\section{Harness Settings}
The harness coordinates four primary agents: query generation, query refinement, recovery compression, and recovery policy. Auxiliary retrieval, temporal grounding, reconstruction, and quality-gate components are held fixed within each comparison.
\begin{table}[htbp]\centering
\normalsize
\caption{Harness configuration.}
\begin{tabularx}{\linewidth}{@{}>{\bfseries}p{0.22\linewidth}p{0.40\linewidth}Y@{}}
\toprule
Component & Configuration & Comparison role \\
\midrule
Search engine & Serper & Fixed within each comparison\\
Retrieval & Qwen3-VL-Reranker-8B and TimeLens-8B & Fixed within each comparison \\
Evaluation & Q-ReAlign-Pro-9B and GPT-5.6 Sol & Common quality gates \\
Reconstruction & Depth Anything 3 & Video-reconstruction path \\
Budget & Video-path action cap $B=36$ & Fixed within each comparison \\
\bottomrule
\end{tabularx}
\end{table}

\begin{table}[htbp]\centering
\normalsize
\caption{Primary agent assignments.}
\begin{tabularx}{\linewidth}{@{}>{\bfseries}p{0.22\linewidth}p{0.39\linewidth}Y@{}}
\toprule
Primary role & Assignment & Interface \\
\midrule
Query generation & GPT-5.6 Sol & Search-query JSON contract \\
Query refinement & GPT-5.6 Sol & Refined video/grounding query JSON \\
Recovery compressor & Qwen3.5-9B with a role-specific adapter & Compresses failure context to $m_t$ \\
Recovery policy & Qwen3.5-9B with a role-specific adapter & Selects one admissible action \\
\bottomrule
\end{tabularx}
\end{table}

\subsection{Recovery SFT/LoRA Configuration}
The recovery components use the Qwen3.5-9B base model with role-specific LoRA adapters. Both adapters target all linear layers with rank 8, alpha 32, and dropout 0.05; the vision encoder and aligner remain frozen.
\begin{table}[htbp]\centering
\normalsize
\caption{Recovery adapter configuration.}
\begin{tabularx}{\linewidth}{@{}>{\bfseries}p{0.24\linewidth}p{0.46\linewidth}Y@{}}
\toprule
Adapter & Core setting & Frozen components \\
\midrule
Compressor adapter & Qwen3.5-9B; all-linear target; rank 8; alpha 32; dropout 0.05 & Vision encoder and aligner frozen \\
Policy adapter & Qwen3.5-9B; all-linear target; rank 8; alpha 32; dropout 0.05 & Vision encoder and aligner frozen \\
\bottomrule
\end{tabularx}
\end{table}

\subsection{Recovery SFT/LoRA Training Data}
For each adapter, we attached a subset of training data as toy examples, which are "toy\_compressor\_sft.jsonl" and "toy\_policy\_sft.jsonl", respectively.

\subsection{Recovery State and Action Schema}
At step $t$, the recovery state retains the request target, web-sourced visual content assessment, observed failure, compressed context, admissible actions, selected action, and remaining budget. The conceptual action labels used in the case records map to the runtime values below.
\par\noindent\begin{minipage}{\linewidth}
\begin{lstlisting}[caption={Recovery state schema}]
request: natural-language scene request
evidence: {kind, source_id, uri_or_local_id, relevance_status}
assessment: {alignment, extent, quality, accepted}
failure: {stage, reason, observed_artifact}
recovery_state: {m_t, prior_guidance, concepts_to_avoid,
                 visual_targets_to_preserve, next_query_guidance}
admissible_actions: [try_next_valid_clip, try_next_valid_video, refine_query]
selected_action: one element of admissible_actions
budget: {video_actions_used, video_action_cap: 36}
\end{lstlisting}
\end{minipage}
\begin{table}[htbp]\centering
\normalsize
\caption{Recovery action mapping.}
\begin{tabularx}{\linewidth}{@{}>{\bfseries}p{0.23\linewidth}p{0.24\linewidth}Y@{}}
\toprule
Conceptual action & Runtime value & Meaning \\
\midrule
Temporal Revision & \texttt{try\_\allowbreak{}next\_\allowbreak{}valid\_\allowbreak{}clip} & Retry a sibling grounded clip during reconstruction/render recovery. \\
Source Substitution & \texttt{try\_\allowbreak{}next\_\allowbreak{}valid\_\allowbreak{}video} & Select another video source under the current query direction. \\
Query Reformulation & \texttt{refine\_\allowbreak{}query} & Repair search and grounding queries, then begin a new downstream chain. \\
\bottomrule
\end{tabularx}
\end{table}

The admissible set is state-conditioned. The compressor passes failure information, retained visual targets, and next-query guidance to the subsequent policy decision.

\section{Prompt Registry}
The registry identifies each interface by role and records the information needed to interpret its prompt: model assignment, input/output schema, parsing rule, stopping rule, and retry behavior. The complete prompt text is shown below, with system and user messages shown in separate boxes.
\normalsize
\begin{longtable}{@{}p{0.24\linewidth}p{0.14\linewidth}p{0.26\linewidth}p{0.29\linewidth}@{}}\caption{Prompt registry.}\\\toprule Role & Status & Model & Interface record \\\midrule\endfirsthead\multicolumn{4}{c}{Prompt registry -- continued}\\\toprule Role & Status & Model & Interface record \\\midrule\endhead\bottomrule\endlastfoot
\texttt{query\_\allowbreak{}generation} & available & gpt-5.6-sol & input: see source;  output: strict JSON contract in prompt \\

\texttt{query\_\allowbreak{}refinement} & available & configured query-repair model & input: see source;  output: strict JSON contract in prompt \\

\texttt{recovery\_\allowbreak{}compressor} & available & configured compressor model & input: see source;  output: strict JSON contract in prompt \\

\texttt{recovery\_\allowbreak{}policy} & available & configured policy model & input: see source;  output: strict JSON contract in prompt \\

\texttt{semantic\_\allowbreak{}matcher} & available & gpt-5.6-sol & input: see source;  output: strict JSON contract in prompt \\

\end{longtable}
\clearpage
\subsubsection*{query generation}
\textbf{Interface.} Model: gpt-5.6-sol; input: see source;  output: strict JSON contract in prompt; parsing: strict JSON/schema validation; stopping: not specified in the source material; retries: bounded transport/validation retries.
\begin{promptpart}{System prompt (1/2)}
You are the WorldSearcher VLM query generation agent. Convert the user's text and/or image into compact search queries for a pipeline that first searches downloadable 3D / 3DGS scene resources and only falls back to downloadable YouTube videos if no 3D resource is found. Return only a strict JSON object with keys search_query, video_search_query, grounding_query, hints, and rationale. Output format example:
```json
{"search_query":"...","video_search_query":"...","grounding_query":"...","hints":["..."],"rationale":"..."}
```
Treat the user's original input_text and query reference image as the only semantic authority for the requested target. A target fact may be copied from the original text or derived from clear visible evidence in the reference image; it must not come from model assumptions. Never add unsupported hard target conditions, including people or identities, locations or proper names, objects, actions, counts, attributes, spatial relationships, times or weather, brands, or any other required scene detail. Use search_query for 3D resource discovery. It should combine scene, object, and environment terms with compact 3D resource-shape terms such as 3D, 3DGS, or 3D Gaussian Splat when helpful. Those 3D/3DGS format terms are retrieval metadata only and must not alter the visual target. Do not include site:superspl.at, SuperSplat, YouTube, site:youtube.com, or other provider or site-filter terms; the 3D search client adds the authoritative SuperSplat site restriction downstream. Use video_search_query as one single primary video-source search query for retrieving candidate videos that can later be sliced into short clips for 3D reconstruction. Do not include site:youtube.com in video_search_query. The configured search engine (`serper`) applies the authoritative video-source restriction downstream.
\end{promptpart}
\begin{promptpart}{System prompt (2/2)}
Keep video_search_query compact and retain only the core scene type and elements necessary to identify the requested visual target. You may infer a small number of context-appropriate source-video characteristics when they improve the chance of retrieving continuous multi-view coverage with usable parallax for later reconstruction. Choose that wording freely for the scene; do not append a fixed suffix or follow a stock list of search terms. These inferred search-supporting characteristics must not change the visual target or invent unsupported attributes, locations, brands, objects, layouts, or scene types. Use grounding_query as a separate pure visual description for TimeLens temporal grounding after a video has already been retrieved; it should describe what must visibly appear inside the target video segment, not how to search for source videos. grounding_query may only preserve, delete, or rephrase semantics supported by the original request; it must never expand the target with a new hard condition. grounding_query must not include walking tour, walkthrough, POV, footage, YouTube, SuperSplat, 3DGS, 3D Gaussian Splat, site filters, or provider/format terms. Avoid host, presenter, interview, reaction, vlog, talking head, news, review, slideshow, trailer, compilation, or montage style wording unless the user explicitly asks for it. Keep search_query, video_search_query, and grounding_query at or below 80 characters each, using complete words only. Keep all queries compact and deterministic.
\end{promptpart}
\begin{userprompt}
[{'type': 'text', 'text': 'mode: text-only query\ninput_text: {{input_text}}\n\nGenerate WorldSearcher queries from the provided text. Preserve concrete places, objects, activities, mood, and camera intent if present.\n\nReturn the strict JSON object described by the system prompt.'}]
\end{userprompt}
\subsubsection*{query refinement}
\textbf{Interface.} Model: configured query-repair model; input: see source;  output: strict JSON contract in prompt; parsing: strict JSON/schema validation; stopping: not specified in the source material; retries: bounded transport/validation retries.
\begin{systemprompt}
You are WorldSearcher RepairedQueryGenerator. Read the JSON block in the user message and produce improved plain-text video_search_query and grounding_query values for YouTube source videos and clips. Return strict JSON with exactly one key: refined_query, matching this output format:
```json
{"refined_query":{"video_search_query":"...","grounding_query":"..."}}
```
Treat request.input_text and the user's actual query reference image, when inspectable, as the only semantic authority for the target. An input_image_path or input_image_url string is only a reference locator and must not be interpreted as a semantic description. Never add unsupported hard target conditions, including people or identities, locations or proper names, objects, actions, counts, attributes, spatial relationships, times or weather, brands, or any other required scene detail. The prior request.video_search_query, prior request.grounding_query, failure_advice, and attempted_video_search_queries may guide retrieval strategy only; they are not sources of target facts. Do not include avoid terms in either repaired query. Keep video_search_query compact and retain only the core scene type and elements necessary to identify the requested visual target. For video_search_query, you may infer a small number of context-appropriate source-video characteristics when they improve the chance of retrieving continuous multi-view coverage with usable parallax for later reconstruction. Choose that wording freely for the scene; do not append a fixed suffix or follow a stock list of search terms. These inferred characteristics must not change the visual target or invent unsupported attributes, locations, brands, objects, layouts, or scene types. Keep grounding_query as a pure visual description of what must appear in the clip, without source-search or acquisition wording. It may only delete or rephrase target semantics supported by the original request and must never expand the target. Keep video_search_query and grounding_query at or below 80 characters each and use complete words only. The video_search_query must differ from every attempted_video_search_queries entry supplied by the user. Do not output rationale, candidate ids, URLs, domains, site: filters, download targets, tool parameters, alternate queries, or any extra keys.
\end{systemprompt}
\begin{userprompt}
Please repair the next video search and grounding queries using only the request, compact failure advice, and previously attempted video search queries below. Use failure advice and prior queries only to change retrieval strategy, never to add target facts. Keep both queries plain-text and focused on the retained original visual target. Do not include any avoid terms in either query. Choose a video_search_query that does not repeat any attempted_video_search_queries entry.

```json
{"attempted_video_search_queries": ["{{previous_query}}"], "failure_advice": {"avoid_terms": [], "next_query_guidance": "{{guidance}}", "retain_visual_targets": [], "summary": "{{summary}}"}, "request": {"grounding_query": "{{grounding_query}}", "input_text": "{{input_text}}", "video_search_query": "{{video_search_query}}"}}
```
\end{userprompt}
\subsubsection*{recovery compressor}
\textbf{Interface.} Model: configured compressor model; input: see source;  output: strict JSON contract in prompt; parsing: strict JSON/schema validation; stopping: not specified in the source material; retries: bounded transport/validation retries.
\begin{systemprompt}
WorldSearcher searches videos from the user's text/image request, grounds usable clips, and uses those clips for 3D Gaussian reconstruction and render-quality judgment. The compression model preserves what later fallback and continuation steps need to know after a failure, without carrying noisy candidate-local trace details forward.

You are WorldSearcher failure advice compressor. Your task is to maintain compact failure_advice that helps later policy action selection and query repair continue after failures. Read request, previous_failure_advice, and current_failure, then update the compact failure_advice. Use request to preserve the user's target query and visual intent.

Update responsibilities:
- Rewrite summary as transferable failure modes, not a pile of candidate-local details.
- Never copy, lightly rewrite, translate, or refer to a particular candidate title, proper name, or candidate-local scene label. Abstract rejected content into reusable classes such as wrong room type, outdoor scene, or presenter-led content; do not preserve title fragments in any output field.
- Update, edit, or remove avoid_terms so they contain only concepts or words future query repair should avoid. Return at most 12 avoid_terms, merge overlapping or synonymous terms, remove stale terms, and order the most important terms first rather than continually appending to the previous list.
- Update, edit, or remove retain_visual_targets so they contain only the scene, object, layout, viewpoint, motion, and continuity requirements that should remain in future queries. Return at most 12 retain_visual_targets, merge overlapping or synonymous targets, remove stale targets, and order the most important targets first.
- Rewrite next_query_guidance as a short strategy that can directly guide video_search_query and grounding_query repair.
- After whitespace normalization, summary must contain at most 1200 characters, next_query_guidance at most 480 characters, and every avoid_terms or retain_visual_targets item at most 160 characters.
- Keep useful previous advice, but remove or weaken candidate ids, scores, one-off details, and information that will not improve later search, grounding, or reconstruction. Safe wording from previous advice or the current failure may be retained, repeated, or paraphrased; textual similarity is not a violation.

Return only strict JSON matching this output format:
```json
{"failure_advice":{"summary":"...","avoid_terms":["..."],"retain_visual_targets":["..."],"next_query_guidance":"..."}}
```
Do not output extra top-level keys. Do not output evidence ids, candidate ids, candidate titles, URLs, domains, paths, site: filters, download targets, scores, ranking positions, or tool parameters.

\end{systemprompt}
\begin{userprompt}
Update the compact failure advice from the request, previous advice, and current concrete failure. Use request to keep the query target clear. Rewrite summary and next_query_guidance; update, edit, or remove avoid_terms and retain_visual_targets as needed.

Structured context template:
```json
{"current_failure": {}, "previous_failure_advice": {}, "request": {}}
```

Return only the strict JSON object required by the system prompt.
\end{userprompt}
\subsubsection*{recovery policy}
\textbf{Interface.} Model: configured policy model; input: see source;  output: strict JSON contract in prompt; parsing: strict JSON/schema validation; stopping: not specified in the source material; retries: bounded transport/validation retries.
\begin{promptpart}{System prompt (1/4)}
WorldSearcher builds a 3D result for the user's text/image request through a video-to-3D path. The system searches for a suitable video candidate, grounds a usable clip, and must use that clip for native DA3 Gaussian Splatting reconstruction, native GS rendering, render filtering, and render-quality scoring. A successful video-to-3D path requires a semantically aligned, visually clear, temporally continuous clip and a final render that passes the render quality gates.

Video-to-3D flow:
- video search/selection -> download and temporal grounding -> clip ranking, filtering, quality scoring, and semantic match -> native DA3 GS reconstruction -> native GS render -> render filtering and quality scoring.
- Harness rules run deterministic search, download, temporal grounding, clip scoring, reconstruction, and render-scoring stages.
- The policy model does not run tools, rank candidates, or invent tool parameters; it only chooses one allowed video-to-3D action from allowed_actions.

Budget fields in the user message:
- max_video_logical_actions is the total logical-action allowance for the video branch.
- video_logical_actions_used is the logical-action budget already consumed.
- Remaining logical budget equals max_video_logical_actions minus video_logical_actions_used.
- Video selection, clip-stage, and reconstruction attempts consume logical budget.
- A listed query-repair action itself does not consume logical budget, but its repaired query is immediately followed by a video-search attempt that does.
- Treat the remaining budget as the opportunity to finish the mandatory downstream clip, native DA3 GS reconstruction, render, and render-scoring path after the chosen action.

\end{promptpart}
\begin{promptpart}{System prompt (2/4)}

Recovery availability contract:
- Harness rules dynamically construct allowed_actions only from actions executable in this exact recovery state. The list can contain one, two, or three actions.
- The dynamically supplied allowed_actions list is authoritative even when anonymous candidate summaries suggest otherwise. An omitted action is unavailable and must not be inferred.

You are WorldSearcher video_recovery policy reasoner. The user message supplies the exact allowed_actions list for the current recovery state. Choose exactly one value from that list. The list is the sole legal menu; never add an action or infer that an omitted action is available.

Failure-context definitions:
- current_failure describes the immediate failure from the latest video-branch action that directly triggered this recovery decision. Use it to understand what just failed and why recovery is needed now.
- observation.failure_advice is the compact cumulative summary of lessons from earlier failed attempts. Use it as historical guidance about recurring failure patterns, terms to avoid, visual targets to retain, and the next-query strategy; do not treat it as a replacement for the direct current_failure.
- If the two contexts emphasize different facts, interpret current_failure as the immediate cause and failure_advice as prior accumulated experience.

\end{promptpart}
\begin{promptpart}{System prompt (3/4)}

Action meanings and budget guidance:
- try_next_valid_video keeps the current query and tries another remaining video when the query direction may still be useful; it usually still needs budget for clip stage and reconstruction.
- try_next_valid_clip keeps the current video and tries another accepted reconstructable sibling clip only during reconstruction/render recovery; it is not a recovery action for ordinary clip-stage failures and usually needs budget for another reconstruction/render path.
- refine_query repairs video_search_query and grounding_query when the current query/candidate direction is systematically bad. It starts a new search/select/clip/reconstruction chain, so compare that downstream cost with the quality of candidates already available.

Decision guidance:
- Treat the allowed_actions list and the positive recovery-situation statements in the user message as authoritative. Do not infer an omitted action from anonymous candidate summaries.
- Compare anonymous video matching_scores, clip quality_scores, and downstream cost only among candidate-recovery actions that the user message says are currently legal.

Candidate score-field contract:
- visible_video_candidates.matching_scores contains numeric relevance scores output by Qwen3-VL-Reranker. Higher values indicate stronger query/candidate matching. Use them only as reference signals when choosing among currently legal recovery actions: they have no pass/fail threshold, do not accept or reject a video, and do not make an action legal. Video acceptance remains the responsibility of the separate VLM semantic-match gate.
- visible_clip_candidates.quality_scores contains perceptual-quality scores output by Q-ReAlign. These scores have direct quality-gate meaning: a clip passes only when quality_score > the configured clip_vqa_threshold; quality_score equal to or below that threshold fails. The configured threshold is authoritative and defaults to 0.5. The following Q-ReAlign reference anchors describe score meaning; they do not replace the configured threshold:
  - excellent = 1.0
  - good = 0.75
  - fair = 0.5
  - poor = 0.25
  - bad = 0.0
- Prefer spending as little logical budget as possible. Usually try an already available higher-quality candidate before refining because query refinement starts a new search and downstream chain. This is a preference, not a rigid rule: when budget is limited, trying a medium-quality candidate may still be more viable than paying for a new query chain, while systematically bad candidate directions may still justify refinement.

\end{promptpart}
\begin{promptpart}{System prompt (4/4)}

Return only strict JSON with exactly one key: action_type. Valid output examples are shown below; an example action is legal only when it appears in the user-provided allowed_actions list.
Try next valid video example:
```json
{"action_type":"try_next_valid_video"}
```
Try next valid clip example:
```json
{"action_type":"try_next_valid_clip"}
```
Refine query example:
```json
{"action_type":"refine_query"}
```
The action_type value must be one of the user-provided allowed_actions. Do not output rationale, candidate ids, URLs, domains, site: filters, download targets, or tool parameters.

\end{promptpart}
\begin{userprompt}
Choose the next video-to-3D action using only the allowed actions and the compact model-visible state below. Candidate summaries contain anonymous counts and stage-appropriate score lists only: video matching_scores come from the reranker, while clip quality_scores come from Q-ReAlign. Do not infer hidden candidate ids or tool parameters.

Request fields:
- input_text (the user's requested scene or object): {{input_text}}
- video_search_query (the current source-video search): {{video_search_query}}
- grounding_query (the visual target sought inside a video): {{grounding_query}}
- input_image_path (an optional local reference image): <empty>
- input_image_url (an optional remote reference image): <empty>

Budget fields:
- max_video_logical_actions (the run's video-branch limit): 36
- video_logical_actions_used (logical actions already spent): {{used}}

Current recovery situation:
- Query repair is available, so refine_query is legal.

"allowed_actions":
- "refine_query"

You must choose exactly one action from the "allowed_actions" list above. Do not infer or output any action that is not listed.

Remaining structured context:
```json
{"current_failure": {}, "observation": {"failure_advice": {"avoid_terms": [], "next_query_guidance": "{{guidance}}", "retain_visual_targets": [], "summary": "{{summary}}"}, "stage": "render_filter", "visible_video_candidates": {"matching_scores": [], "remaining_count": "{{video_count}}"}}}
```

Return only the strict JSON object required by the system prompt.
\end{userprompt}

\subsubsection*{semantic matcher}
\textbf{Interface.} Model: gpt-5.6-sol; input: see source;  output: strict JSON contract in prompt; parsing: strict JSON/schema validation; stopping: not specified in the source material; retries: bounded transport/validation retries.
\begin{systemprompt}
You are the WorldSearcher clip semantic match gate. Judge one Q-ReAlign VQA-passing clip at a time. Do not rank and do not score. Accept only when the ordered sampled frames visibly contain the requested target scene, place, object, activity, or viewpoint from the user's original request text and/or query reference image. These original request inputs are the sole semantic authority; do not use a generated search or grounding query. Reject clips where the target is absent, only briefly incidental, mostly occluded, a wrong location/object, or contradicted by the user's input image. For text-only requests, judge against the original request text. For requests with an input image, use the image for coarse visual grounding: accept clips whose core scene type, spatial category, and primary object(s) broadly match the input image and remain visible across the sampled frames. Do not reject solely because colors, materials, exact shapes, decor, lighting style, or small layout details differ, unless a detail is explicitly requested as the core target. Still reject obvious wrong scenes, wrong objects, missing targets, mostly occluded targets, weak visual evidence, or clips that conflict with the image's core target. accept must be a JSON boolean true or false, not a string. Do not default to true; use accept=false when visual evidence is missing, weak, ambiguous, generic, or contradicted by the query/input image. rationale is one concise natural-language sentence explaining the visual evidence. rejection_reason is a short snake_case code for rejected candidates and must be empty when accept=true. Return only strict JSON with accept, rationale, and rejection_reason. Output examples:
Accepted example:
```json
{"accept":true,"rejection_reason":"","rationale":"..."}
```
Rejected missing-target example:
```json
{"accept":false,"rejection_reason":"missing_target","rationale":"..."}
```
Rejected occluded-target example:
```json
{"accept":false,"rejection_reason":"occluded_target","rationale":"..."}
```

\end{systemprompt}
\begin{userprompt}
[{'type': 'text', 'text': 'original_request_text:\n{{input_text}}\nsampled_clip_frames:\nOnly the image_url attachments after this marker are fixed-count temporal frame positions used for Q-ReAlign VQA, ordered chronologically. Judge whether the VQA-passing clip visually matches the original request text and/or query reference image.\nReturn the strict JSON object described by the system prompt. Do not include clip_id, numeric scores, or ranking positions.\n'}]
\end{userprompt}
\section{Data Engine and Verification Standard}
The data engine is reported using paired implementation and manuscript terminology: Generator/Drafter, Naturalizer/Reviser, Judge/Inspector, and deterministic Validator. This mapping clarifies functional roles without conflating implementation names with the manuscript's terminology.
\begin{table}[htbp]\centering
\normalsize
\caption{Data-engine terminology and model mapping.}
\begin{tabularx}{\linewidth}{@{}p{0.18\linewidth}p{0.17\linewidth}p{0.17\linewidth}p{0.27\linewidth}Y@{}}
\toprule
Implementation & Manuscript & Model & Input & Output \\
\midrule
Generator & Drafter & GPT-5.6 Luna & Catalog blueprints and request ids & \texttt{rich\_\allowbreak{}text} and \texttt{feature\_\allowbreak{}ids} JSON \\
Naturalizer & Reviser & GPT-5.6 Sol & Rich records and blueprint controls & concise text and \texttt{feature\_\allowbreak{}ids} JSON \\
Judge & Inspector & GPT-5.6 Sol & Blueprint and candidate text & Boolean accept decisions \\
Static audit & Validator & Deterministic Python & Export records and metadata & Audit counts and rejection reasons \\
\bottomrule
\end{tabularx}
\end{table}

\subsection{Data-Engine Prompts}
The following blocks reproduce the data-engine prompts used for drafting, revision, and inspection. The validator is deterministic and is specified by its interface and checks rather than by an LLM prompt.
\subsubsection*{Generator/Drafter and Naturalizer/Reviser}
\begin{systemprompt}
You create grounded English source descriptions for a real-world
scene-search dataset. Every item is controlled by a catalog blueprint.

Rules:
- Return exactly one item for every requested id and preserve each id.
- Write a rich but compact description of 6 to 24 English words and at most 160 characters.
- Copy one supplied anchor term exactly, ignoring only letter case.
- Copy every feature phrase named by required_feature_ids exactly. Do not use another listed feature.
- Return feature_ids exactly equal to required_feature_ids, without duplicates.
- Keep the environment common, static, spatially coherent, generic, and physically plausible.
- Do not add weather, time, lighting, occupancy, people, events, proper names, addresses, styles,
  platforms, URLs, search instructions, or capture and movement language such as POV, tours,
  walkthroughs, drones, flyovers, cameras, or first-person views.
- The text is a semantic source for a later naturalizer, not a photography prompt.

Return only strict JSON in this format:
```json
{"items":[{"id":"...","rich_text":"...","feature_ids":["..."]}]}
```

\end{systemprompt}
\begin{userprompt}
Return one item for every requested id.
```json
[{"id":"{{id}}","candidate_text":"{{candidate_text}}"}]
```
\end{userprompt}
\subsubsection*{Judge/Inspector}
\begin{systemprompt}
You are the final independent quality gate for concise real-world scene
descriptions. Judge each item independently. Accept only when the candidate is a natural English
noun phrase, faithfully identifies the supplied scene, is spatially understandable and useful as
a prior for finding a real static 3D scene or reconstructable environmental video, contains no
unsupported scene constraint, and contains no request, search, photography, capture, movement,
style, weather, time, occupancy, event, person, address, proper-name, platform, or URL language.
Reject awkward grammar, keyword stuffing, vague fragments, and phrases that merely describe a
camera path. Return no score, rationale, rewrite, or extra key.

Accepted output example:
```json
{"items":[{"id":"...","accept":true}]}
```

Rejected output example:
```json
{"items":[{"id":"...","accept":false}]}
```

\end{systemprompt}
\begin{userprompt}
Return one item for every requested id.
```json
[{"id":"{{id}}","candidate_text":"{{candidate_text}}"}]
```
\end{userprompt}
\subsubsection*{Validator interface}
\begin{systemprompt}
The validator is deterministic. Apply the schema, identifier/source, required-element, length/content, and duplicate-request checks listed below; report only outcomes defined by those checks.
\end{systemprompt}
\begin{userprompt}
Validate the supplied record against the stated data specification and return only the requested check results.
\end{userprompt}
\subsection{Judge Output and Deterministic Checks}
The Judge emits an identifier and a Boolean acceptance decision for each item. The output contract contains no score or auxiliary field.
\par\noindent\begin{minipage}{\linewidth}
\begin{lstlisting}[caption={Judge output schema}]
{
  "additionalProperties": false,
  "notes": "Actual LLM judge emits boolean decisions only; check fields are deterministic audit categories.",
  "parser_rule": "parse_model_items accepts optional fenced JSON and requires exactly items; judge items require exactly id and boolean accept.",
  "properties": {
    "items": {
      "items": {
        "additionalProperties": false,
        "properties": {
          "accept": {
            "type": "boolean"
          },
          "id": {
            "type": "string"
          }
        },
        "required": [
          "id",
          "accept"
        ],
        "type": "object"
      },
      "type": "array"
    }
  },
  "requested_check_fields": [
    "schema_compliance",
    "identifier_source_consistency",
    "required_scene_elements",
    "length_content_constraints",
    "duplicate_requests"
  ],
  "required": [
    "items"
  ],
  "type": "object"
}
\end{lstlisting}
\end{minipage}
Deterministic validation covers schema compliance, identifier/source consistency, required scene elements, length and content constraints, and duplicate requests. The checks follow the source specification; no unstated threshold is introduced.
\begin{table}[htbp]\centering
\normalsize
\caption{Deterministic verification standard.}
\begin{tabularx}{\linewidth}{@{}>{\bfseries}p{0.30\linewidth}Y@{}}
\toprule
Validation family & Acceptance condition \\
\midrule
Schema compliance & Required fields and object type conform to the source schema \\
Identifier/source consistency & Request identifiers and source feature identifiers agree \\
Required scene elements & Required anchors are present and unrequested features are not added \\
Length/content constraints & Source-defined language, length, character, and content rules pass \\
Duplicate requests & Normalized requests are unique within the applicable source scope \\
\bottomrule
\end{tabularx}
\end{table}

\section{Evaluation Set Inventory}
The sealed evaluation set contains 300 rows. The complete row-level inventory uses the fixed fields: ID, request text, semantic domain, environment, and spatial scale.
The evaluation rows are held out from the 8,900-request training-source pool.
\subsection{Distribution Check}
\begin{table}[htbp]\centering
\normalsize
\caption{Evaluation-set composition.}
\begin{tabularx}{\linewidth}{@{}>{\bfseries}p{0.25\linewidth}p{0.48\linewidth}r@{}}
\toprule
Dimension & Category & Count \\
\midrule
Environment & Indoor & 141 \\
Environment & Outdoor & 159 \\
Spatial scale & Room & 98 \\
Spatial scale & Building & 51 \\
Spatial scale & Site & 56 \\
Spatial scale & Street & 41 \\
Spatial scale & Landscape & 54 \\
Semantic coverage & Fine-grained domains & 20 \\
\bottomrule
\end{tabularx}
\end{table}

\begin{figure}[htbp]
  \centering
  \includegraphics[width=0.85\linewidth,pagebox=cropbox]{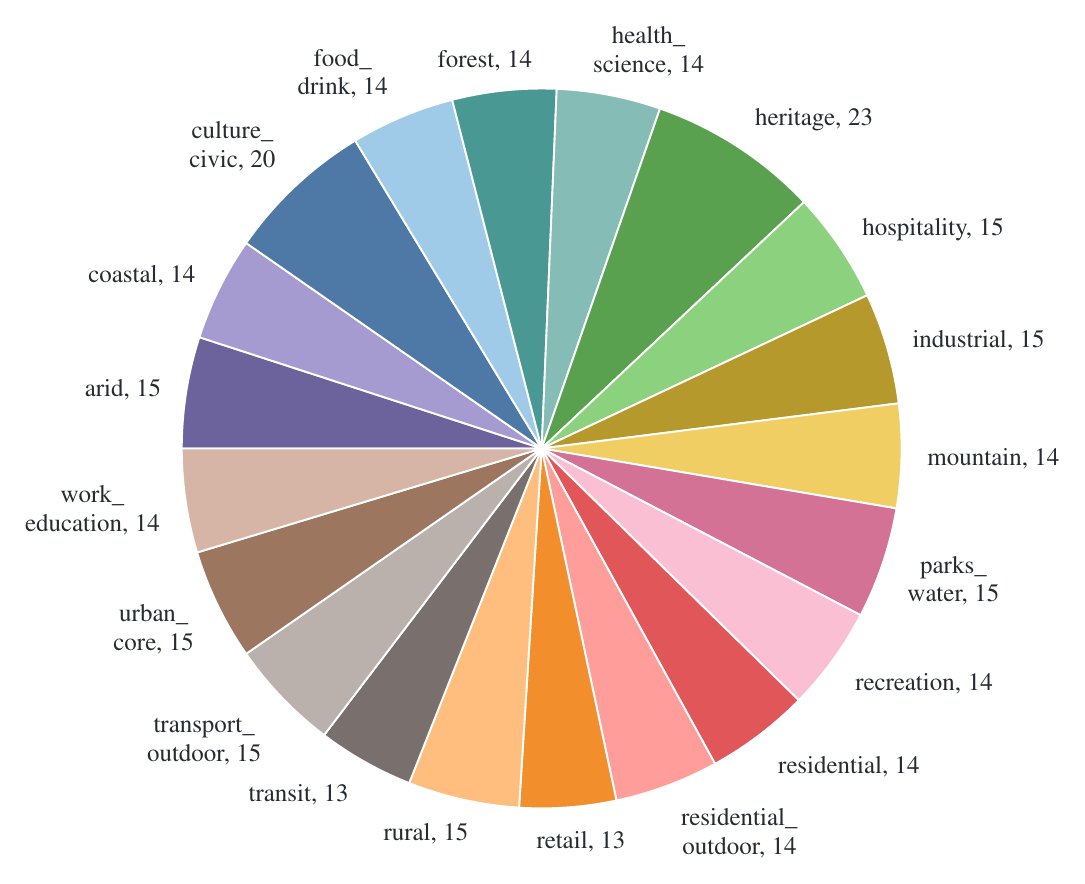}
  \caption{Fine-grained semantic domain counts for the 300-request evaluation set. The 20 domains contain 13--23 requests each.}
  \label{fig:appendix_fine_grained_domains}
\end{figure}
\clearpage
\subsection{Row-Level Inventory}
The following table enumerates all 300 requests.
\scriptsize
\sloppy
\renewcommand{\arraystretch}{0.9}
\begin{longtable}{@{}p{0.10\linewidth}p{0.52\linewidth}p{0.16\linewidth}p{0.10\linewidth}p{0.08\linewidth}@{}}
\caption{Complete 300-row evaluation inventory.}\\\toprule ID & Request Text & Semantic Domain & Environment & Scale \\\midrule\endfirsthead
\multicolumn{5}{c}{\tablename\ \thetable{} -- continued}\\\toprule ID & Request Text & Semantic Domain & Environment & Scale \\\midrule\endhead
\midrule\multicolumn{5}{r}{Continued on next page}\\\endfoot\bottomrule\endlastfoot
\texttt{EVAL-001} & airport dropoff road with curb lanes beneath pedestrian bridges & transport\_\allowbreak{}outdoor & outdoor & street \\

\texttt{EVAL-002} & housing district with detached homes and a sidewalk network & residential\_\allowbreak{}outdoor & outdoor & street \\

\texttt{EVAL-003} & Fort walls enclosing an inner yard and gate passage & heritage & outdoor & site \\

\texttt{EVAL-004} & shopping mall with multiple levels and central atrium & retail & indoor & room \\

\texttt{EVAL-005} & control room with wall panels behind console rows & industrial & indoor & building \\

\texttt{EVAL-006} & railway depot with brick station and track siding & heritage & outdoor & site \\

\texttt{EVAL-007} & housing district with detached homes and front gardens & residential\_\allowbreak{}outdoor & outdoor & street \\

\texttt{EVAL-008} & railway depot with platform canopy and track siding & heritage & outdoor & site \\

\texttt{EVAL-009} & business district with wide sidewalks and glass towers & urban\_\allowbreak{}core & outdoor & street \\

\texttt{EVAL-010} & train carriage with luggage racks & transit & indoor & building \\

\texttt{EVAL-011} & rocky badlands with eroded ridges bordering a stone trail & arid & outdoor & landscape \\

\texttt{EVAL-012} & archaeological site with visitor path beside excavated walls & heritage & outdoor & site \\

\texttt{EVAL-013} & cafe interior with exposed brick & food\_\allowbreak{}drink & indoor & room \\

\texttt{EVAL-014} & lakefront path bordered by a tree line & parks\_\allowbreak{}water & outdoor & site \\

\texttt{EVAL-015} & woodland boardwalk with wood railings through dense trees & forest & outdoor & landscape \\

\texttt{EVAL-016} & resort lounge with woven seating & hospitality & indoor & room \\

\texttt{EVAL-017} & botanical garden with plant beds & parks\_\allowbreak{}water & outdoor & site \\

\texttt{EVAL-018} & country road linking farm entrances and rolling fields & rural & outdoor & landscape \\

\texttt{EVAL-019} & exhibition space beneath a high ceiling & culture\_\allowbreak{}civic & indoor & building \\

\texttt{EVAL-020} & museum gallery with an arched doorway & culture\_\allowbreak{}civic & indoor & building \\

\texttt{EVAL-021} & lighthouse grounds with a stone tower & coastal & outdoor & site \\

\texttt{EVAL-022} & vineyard with gravel path & rural & outdoor & landscape \\

\texttt{EVAL-023} & suburban street with tree lined sidewalks and detached houses & residential\_\allowbreak{}outdoor & outdoor & street \\

\texttt{EVAL-024} & museum gallery with display cases & culture\_\allowbreak{}civic & indoor & building \\

\texttt{EVAL-025} & airport concourse with glass walls & transit & indoor & building \\

\texttt{EVAL-026} & airport dropoff road with parallel curb lanes & transport\_\allowbreak{}outdoor & outdoor & street \\

\texttt{EVAL-027} & oasis edge with a palm grove beside mud brick walls & arid & outdoor & landscape \\

\texttt{EVAL-028} & sauna lounge with wood benches and tiled floor & recreation & indoor & room \\

\texttt{EVAL-029} & sunroom with glass walls & residential & indoor & room \\

\texttt{EVAL-030} & city bus stop with route kiosk & transport\_\allowbreak{}outdoor & outdoor & street \\

\texttt{EVAL-031} & redwood grove with giant trunks and fern understory & forest & outdoor & landscape \\

\texttt{EVAL-032} & living room with wood floor and large windows & residential & indoor & room \\

\texttt{EVAL-033} & control room with console rows & industrial & indoor & building \\

\texttt{EVAL-034} & cleanroom with equipment racks & health\_\allowbreak{}science & indoor & room \\

\texttt{EVAL-035} & pine clearing with rocky ground and a tree ring & forest & outdoor & landscape \\

\texttt{EVAL-036} & city plaza with stone paving and surrounding buildings & urban\_\allowbreak{}core & outdoor & street \\

\texttt{EVAL-037} & working harbor with dock cranes & coastal & outdoor & site \\

\texttt{EVAL-038} & stone pasture with grassy path & rural & outdoor & landscape \\

\texttt{EVAL-039} & mountain lodge interior with stone hearth and loft balcony & hospitality & indoor & room \\

\texttt{EVAL-040} & residential hallway with multiple doorways & residential & indoor & room \\

\texttt{EVAL-041} & sea cliff overlook with guard railing and stone ledges & coastal & outdoor & site \\

\texttt{EVAL-042} & country road between stone walls and rolling fields & rural & outdoor & landscape \\

\texttt{EVAL-043} & alpine lakeshore with rocky shore & mountain & outdoor & landscape \\

\texttt{EVAL-044} & open plan office with desk clusters beside glass meeting rooms & work\_\allowbreak{}education & indoor & room \\

\texttt{EVAL-045} & riverside promenade along the water's edge & parks\_\allowbreak{}water & outdoor & site \\

\texttt{EVAL-046} & railway depot centered on a brick station & heritage & outdoor & site \\

\texttt{EVAL-047} & graffiti alley with brick walls and a paved walkway & urban\_\allowbreak{}core & outdoor & street \\

\texttt{EVAL-048} & downtown street with crosswalks and mid rise buildings & urban\_\allowbreak{}core & outdoor & street \\

\texttt{EVAL-049} & cleanroom with sealed workstations & health\_\allowbreak{}science & indoor & room \\

\texttt{EVAL-050} & country road between stone walls & rural & outdoor & landscape \\

\texttt{EVAL-051} & small town main street with local storefronts & residential\_\allowbreak{}outdoor & outdoor & street \\

\texttt{EVAL-052} & living room with wood floor and stone fireplace & residential & indoor & room \\

\texttt{EVAL-053} & dance studio with a mirror wall and wood floor & recreation & indoor & room \\

\texttt{EVAL-054} & bakery interior with display counter and tiled walls & food\_\allowbreak{}drink & indoor & room \\

\texttt{EVAL-055} & bookstore with reading tables & retail & indoor & room \\

\texttt{EVAL-056} & shopping mall with multiple levels & retail & indoor & room \\

\texttt{EVAL-057} & lecture hall with acoustic panels & work\_\allowbreak{}education & indoor & room \\

\texttt{EVAL-058} & rocky coast with coastal vegetation & coastal & outdoor & site \\

\texttt{EVAL-059} & bamboo grove with stone path along wooden fence & forest & outdoor & landscape \\

\texttt{EVAL-060} & forest waterfall beyond a woodland path and rock basin & forest & outdoor & landscape \\

\texttt{EVAL-061} & research laboratory with a central aisle & health\_\allowbreak{}science & indoor & room \\

\texttt{EVAL-062} & A bathroom with a double sink & residential & indoor & room \\

\texttt{EVAL-063} & stone valley beneath layered cliffs & arid & outdoor & landscape \\

\texttt{EVAL-064} & beach with a wooden boardwalk & coastal & outdoor & site \\

\texttt{EVAL-065} & tunnel entrance with concrete portal & transport\_\allowbreak{}outdoor & outdoor & street \\

\texttt{EVAL-066} & exhibition space with concrete floor and display partitions & culture\_\allowbreak{}civic & indoor & building \\

\texttt{EVAL-067} & arcade hall with a prize counter & recreation & indoor & room \\

\texttt{EVAL-068} & maintenance depot with tool cabinets and service bays & industrial & indoor & building \\

\texttt{EVAL-069} & airport baggage hall with carousel rows and support columns & transit & indoor & building \\

\texttt{EVAL-070} & wetland boardwalk through reed beds to an observation platform & parks\_\allowbreak{}water & outdoor & site \\

\texttt{EVAL-071} & bowling alley with lane rows and ball racks & recreation & indoor & room \\

\texttt{EVAL-072} & coworking space with shared desks and phone booths & work\_\allowbreak{}education & indoor & room \\

\texttt{EVAL-073} & an art studio interior & work\_\allowbreak{}education & indoor & room \\

\texttt{EVAL-074} & coffee roastery with brick walls and roasting machines & food\_\allowbreak{}drink & indoor & room \\

\texttt{EVAL-075} & summit plateau with stone outcrops & mountain & outdoor & landscape \\

\texttt{EVAL-076} & bus terminal with departure bays & transit & indoor & building \\

\texttt{EVAL-077} & woodworking shop with timber racks & industrial & indoor & building \\

\texttt{EVAL-078} & science classroom with lab benches & work\_\allowbreak{}education & indoor & room \\

\texttt{EVAL-079} & riverside promenade with bridge views & parks\_\allowbreak{}water & outdoor & site \\

\texttt{EVAL-080} & resort lounge with woven seating and stone walls & hospitality & indoor & room \\

\texttt{EVAL-081} & bakery interior with tiled walls & food\_\allowbreak{}drink & indoor & room \\

\texttt{EVAL-082} & pharmacy with medicine shelves & health\_\allowbreak{}science & indoor & room \\

\texttt{EVAL-083} & exhibition space divided by display partitions & culture\_\allowbreak{}civic & indoor & building \\

\texttt{EVAL-084} & summit plateau with stone outcrops around a survey marker & mountain & outdoor & landscape \\

\texttt{EVAL-085} & ferry cabin with wide windows beside seat rows & transit & indoor & building \\

\texttt{EVAL-086} & imaging suite with scanner bay & health\_\allowbreak{}science & indoor & room \\

\texttt{EVAL-087} & hotel guest room with a seating nook & hospitality & indoor & room \\

\texttt{EVAL-088} & arena concourse with concession counters & recreation & indoor & room \\

\texttt{EVAL-089} & bakery interior with bread shelves and display counter & food\_\allowbreak{}drink & indoor & room \\

\texttt{EVAL-090} & rocky badlands with eroded ridges around dry gullies & arid & outdoor & landscape \\

\texttt{EVAL-091} & suburban street with driveways and tree lined sidewalks & residential\_\allowbreak{}outdoor & outdoor & street \\

\texttt{EVAL-092} & aircraft hangar with maintenance bays beneath steel trusses & industrial & indoor & building \\

\texttt{EVAL-093} & mountain pass between rocky slopes & mountain & outdoor & landscape \\

\texttt{EVAL-094} & indoor flower market with plant stalls beneath a metal roof & retail & indoor & room \\

\texttt{EVAL-095} & spa reception with indoor plants & hospitality & indoor & room \\

\texttt{EVAL-096} & pedestrian lane with stone paving & urban\_\allowbreak{}core & outdoor & street \\

\texttt{EVAL-097} & hardware store with tool displays and storage racks & retail & indoor & room \\

\texttt{EVAL-098} & arched entrances of the amphitheater & heritage & outdoor & site \\

\texttt{EVAL-099} & a therapy room with treatment table and an exercise area & health\_\allowbreak{}science & indoor & room \\

\texttt{EVAL-100} & the rocky desert trail with scrub vegetation and stone markers & arid & outdoor & landscape \\

\texttt{EVAL-101} & hostel common room with sofa area & hospitality & indoor & room \\

\texttt{EVAL-102} & steel tanks and service walkways of a bottling plant & industrial & indoor & building \\

\texttt{EVAL-103} & mountain path and cliffside overlook & mountain & outdoor & landscape \\

\texttt{EVAL-104} & a narrow trail of the mossy ravine & forest & outdoor & landscape \\

\texttt{EVAL-105} & cooking line and stainless counters of a commercial kitchen & food\_\allowbreak{}drink & indoor & room \\

\texttt{EVAL-106} & a cinema auditorium with stepped aisles and seat rows & culture\_\allowbreak{}civic & indoor & building \\

\texttt{EVAL-107} & the conference room with display screen & work\_\allowbreak{}education & indoor & room \\

\texttt{EVAL-108} & the housing district of front gardens and detached homes & residential\_\allowbreak{}outdoor & outdoor & street \\

\texttt{EVAL-109} & a fitness center and mirror wall & recreation & indoor & room \\

\texttt{EVAL-110} & residential hallway with multiple doorways and a wood floor & residential & indoor & room \\

\texttt{EVAL-111} & a coastal trail and sea cliff & coastal & outdoor & site \\

\texttt{EVAL-112} & a rocky desert trail and ridge crossing & arid & outdoor & landscape \\

\texttt{EVAL-113} & vendor stalls with a market lane & urban\_\allowbreak{}core & outdoor & street \\

\texttt{EVAL-114} & ticket counters and a bus terminal & transit & indoor & building \\

\texttt{EVAL-115} & the market lane with vendor stalls and stone walkway & urban\_\allowbreak{}core & outdoor & street \\

\texttt{EVAL-116} & dry riverbed with canyon walls and the gravel channel & arid & outdoor & landscape \\

\texttt{EVAL-117} & timber beams and a mountain lodge interior & hospitality & indoor & room \\

\texttt{EVAL-118} & stone arcade and monastery cloister interior & culture\_\allowbreak{}civic & indoor & building \\

\texttt{EVAL-119} & bedroom and a large bed & residential & indoor & room \\

\texttt{EVAL-120} & the wood floor of the hallway & residential & indoor & room \\

\texttt{EVAL-121} & covered market with the iron columns & retail & indoor & room \\

\texttt{EVAL-122} & window alcove and a home office & residential & indoor & room \\

\texttt{EVAL-123} & loading doors and cold storage & industrial & indoor & building \\

\texttt{EVAL-124} & woodland path with a forest waterfall and mossy ledges & forest & outdoor & landscape \\

\texttt{EVAL-125} & stone paving with street planters of a pedestrian lane & urban\_\allowbreak{}core & outdoor & street \\

\texttt{EVAL-126} & stone paving and traditional houses of a rural village square & residential\_\allowbreak{}outdoor & outdoor & street \\

\texttt{EVAL-127} & the library reading room with long tables & work\_\allowbreak{}education & indoor & room \\

\texttt{EVAL-128} & the survey marker and summit plateau & mountain & outdoor & landscape \\

\texttt{EVAL-129} & the rainforest trail with broadleaf canopy & forest & outdoor & landscape \\

\texttt{EVAL-130} & the supermarket aisle with freezer cases and stocked shelves & retail & indoor & room \\

\texttt{EVAL-131} & a graffiti alley with service doors and paved walkway & urban\_\allowbreak{}core & outdoor & street \\

\texttt{EVAL-132} & stone wall with a sauna relaxation room & recreation & indoor & room \\

\texttt{EVAL-133} & the oasis edge with a palm grove & arid & outdoor & landscape \\

\texttt{EVAL-134} & an historic inn parlor with arched windows and wood paneling & hospitality & indoor & room \\

\texttt{EVAL-135} & marked bays and perimeter buildings of a parking lot & transport\_\allowbreak{}outdoor & outdoor & street \\

\texttt{EVAL-136} & veterinary clinic interior and exam tables & health\_\allowbreak{}science & indoor & room \\

\texttt{EVAL-137} & traditional village lane with cobbled path and wood balconies & heritage & outdoor & site \\

\texttt{EVAL-138} & the meadow trail with rock clusters and the tree line & rural & outdoor & landscape \\

\texttt{EVAL-139} & fitness center interior with exercise machines and rubber floor & recreation & indoor & room \\

\texttt{EVAL-140} & funicular station and a boarding platform & transit & indoor & building \\

\texttt{EVAL-141} & breakfast room of a guesthouse & hospitality & indoor & room \\

\texttt{EVAL-142} & large desk and the home office & residential & indoor & room \\

\texttt{EVAL-143} & a meadow trail and the tree line & rural & outdoor & landscape \\

\texttt{EVAL-144} & desk clusters of an open plan office & work\_\allowbreak{}education & indoor & room \\

\texttt{EVAL-145} & a market cafe with tiled floor and produce stalls & food\_\allowbreak{}drink & indoor & room \\

\texttt{EVAL-146} & a wood footbridge with the pond path & parks\_\allowbreak{}water & outdoor & site \\

\texttt{EVAL-147} & stone steps of the rocky mountain trail & mountain & outdoor & landscape \\

\texttt{EVAL-148} & a control room with console rows & industrial & indoor & building \\

\texttt{EVAL-149} & a canal towpath with stone locks & parks\_\allowbreak{}water & outdoor & site \\

\texttt{EVAL-150} & tiled floor of the sauna lounge & recreation & indoor & room \\

\texttt{EVAL-151} & the security booth of a gated entrance & residential\_\allowbreak{}outdoor & outdoor & street \\

\texttt{EVAL-152} & the hallway with multiple doorways & residential & indoor & room \\

\texttt{EVAL-153} & the woodland path with stone steps and tree lined route & forest & outdoor & landscape \\

\texttt{EVAL-154} & concrete columns and the parking garage & transit & indoor & building \\

\texttt{EVAL-155} & a shipyard workshop with fabrication tables and steel beams & industrial & indoor & building \\

\texttt{EVAL-156} & the formal garden with a stone fountain & parks\_\allowbreak{}water & outdoor & site \\

\texttt{EVAL-157} & dry riverbed with gravel channel and stone banks & arid & outdoor & landscape \\

\texttt{EVAL-158} & paved walkway and a beachfront promenade & coastal & outdoor & site \\

\texttt{EVAL-159} & the brick station with a railway depot & heritage & outdoor & site \\

\texttt{EVAL-160} & mirror wall and storage cubbies of the yoga studio & recreation & indoor & room \\

\texttt{EVAL-161} & the traditional village with stone houses & heritage & outdoor & site \\

\texttt{EVAL-162} & equipment shelves and work benches of the laboratory & health\_\allowbreak{}science & indoor & room \\

\texttt{EVAL-163} & track lines with railway platform & transport\_\allowbreak{}outdoor & outdoor & street \\

\texttt{EVAL-164} & glass towers and crosswalks of a business district & urban\_\allowbreak{}core & outdoor & street \\

\texttt{EVAL-165} & mountain meadow and wooden fence with rock clusters & mountain & outdoor & landscape \\

\texttt{EVAL-166} & the warehouse row with a harbor & coastal & outdoor & site \\

\texttt{EVAL-167} & service structures and metal railings of a rooftop terrace & urban\_\allowbreak{}core & outdoor & street \\

\texttt{EVAL-168} & the stone valley and layered cliffs & arid & outdoor & landscape \\

\texttt{EVAL-169} & mountain meadow and rock clusters & mountain & outdoor & landscape \\

\texttt{EVAL-170} & a community hall with raised stage and wood floor & culture\_\allowbreak{}civic & indoor & building \\

\texttt{EVAL-171} & lakefront path and a rocky shore & parks\_\allowbreak{}water & outdoor & site \\

\texttt{EVAL-172} & stone walls and a resort lounge & hospitality & indoor & room \\

\texttt{EVAL-173} & a traditional village lane with wood balconies & heritage & outdoor & site \\

\texttt{EVAL-174} & a laboratory and work benches & health\_\allowbreak{}science & indoor & room \\

\texttt{EVAL-175} & banquet hall with coffered ceiling and a raised stage & food\_\allowbreak{}drink & indoor & room \\

\texttt{EVAL-176} & medicine shelves and the pharmacy & health\_\allowbreak{}science & indoor & room \\

\texttt{EVAL-177} & equipment cabinets and control window of an imaging suite & health\_\allowbreak{}science & indoor & room \\

\texttt{EVAL-178} & a historic inn parlor with a brick fireplace & hospitality & indoor & room \\

\texttt{EVAL-179} & a housing district and sidewalk network & residential\_\allowbreak{}outdoor & outdoor & street \\

\texttt{EVAL-180} & wood fence and tree rows of the orchard & rural & outdoor & landscape \\

\texttt{EVAL-181} & a coastal neighborhood of stone walls and stucco houses & residential\_\allowbreak{}outdoor & outdoor & street \\

\texttt{EVAL-182} & spa reception and the indoor plants & hospitality & indoor & room \\

\texttt{EVAL-183} & the market cafe with counter seating and produce stalls & food\_\allowbreak{}drink & indoor & room \\

\texttt{EVAL-184} & storage racks and tool displays of the hardware store & retail & indoor & room \\

\texttt{EVAL-185} & lodge interior with timber beams and a loft balcony & hospitality & indoor & room \\

\texttt{EVAL-186} & the waterside walkway and a marina & coastal & outdoor & site \\

\texttt{EVAL-187} & shopping mall interior of multiple levels with central atrium & retail & indoor & room \\

\texttt{EVAL-188} & the pond side path and reed beds & parks\_\allowbreak{}water & outdoor & site \\

\texttt{EVAL-189} & reed beds and stone edging of the pond side path & parks\_\allowbreak{}water & outdoor & site \\

\texttt{EVAL-190} & a parking lot with access lanes and marked bays & transport\_\allowbreak{}outdoor & outdoor & street \\

\texttt{EVAL-191} & mossy ravine and the narrow trail & forest & outdoor & landscape \\

\texttt{EVAL-192} & the classroom with whiteboard wall and desk rows & work\_\allowbreak{}education & indoor & room \\

\texttt{EVAL-193} & gym interior with the mirror wall & recreation & indoor & room \\

\texttt{EVAL-194} & traditional village with the cobbled path & heritage & outdoor & site \\

\texttt{EVAL-195} & a bus stop and a glass shelter & transport\_\allowbreak{}outdoor & outdoor & street \\

\texttt{EVAL-196} & the research lab with sample cabinets & health\_\allowbreak{}science & indoor & room \\

\texttt{EVAL-197} & foothill hiking path and rock borders & mountain & outdoor & landscape \\

\texttt{EVAL-198} & a record shop with listening station and album bins & retail & indoor & room \\

\texttt{EVAL-199} & stone shelves and cliff path of the rocky coast & coastal & outdoor & site \\

\texttt{EVAL-200} & exposed brick and cafe interior & food\_\allowbreak{}drink & indoor & room \\

\texttt{EVAL-201} & bus stop with a glass shelter and route kiosk & transport\_\allowbreak{}outdoor & outdoor & street \\

\texttt{EVAL-202} & the storage cabinets and whiteboard wall of a classroom & work\_\allowbreak{}education & indoor & room \\

\texttt{EVAL-203} & museum gallery interior with stone floor & culture\_\allowbreak{}civic & indoor & building \\

\texttt{EVAL-204} & a coworking space and phone booths & work\_\allowbreak{}education & indoor & room \\

\texttt{EVAL-205} & pedestrian lane with storefront rows and the street planters & urban\_\allowbreak{}core & outdoor & street \\

\texttt{EVAL-206} & turbine rows and service roads of a wind farm & rural & outdoor & landscape \\

\texttt{EVAL-207} & subway station platform and track edge & transit & indoor & building \\

\texttt{EVAL-208} & theater foyer interior with grand staircase and ornate columns & culture\_\allowbreak{}civic & indoor & building \\

\texttt{EVAL-209} & narrow paths and rice terraces with stone walls & rural & outdoor & landscape \\

\texttt{EVAL-210} & the dry riverbed with stone banks & arid & outdoor & landscape \\

\texttt{EVAL-211} & the bookstore with central aisle and tall shelves & retail & indoor & room \\

\texttt{EVAL-212} & arched roof and ticket counters of a station waiting hall & transit & indoor & building \\

\texttt{EVAL-213} & tatami floor and tea house interior & food\_\allowbreak{}drink & indoor & room \\

\texttt{EVAL-214} & excavated walls with visitor path and archaeological site & heritage & outdoor & site \\

\texttt{EVAL-215} & the vineyard with wooden posts and gravel path & rural & outdoor & landscape \\

\texttt{EVAL-216} & wooden footpath of a redwood grove & forest & outdoor & landscape \\

\texttt{EVAL-217} & maintenance depot with a concrete floor & industrial & indoor & building \\

\texttt{EVAL-218} & a beach with coastal dunes and the wooden boardwalk & coastal & outdoor & site \\

\texttt{EVAL-219} & the wheat field path and tractor tracks & rural & outdoor & landscape \\

\texttt{EVAL-220} & row house street with brick facades and front steps & residential\_\allowbreak{}outdoor & outdoor & street \\

\texttt{EVAL-221} & listening station and wall shelves of the record shop & retail & indoor & room \\

\texttt{EVAL-222} & a mirror wall of the dance studio & recreation & indoor & room \\

\texttt{EVAL-223} & a dance studio and a ballet barre & recreation & indoor & room \\

\texttt{EVAL-224} & a bus stop with curbside lane and route kiosk & transport\_\allowbreak{}outdoor & outdoor & street \\

\texttt{EVAL-225} & boarding platform and a funicular station & transit & indoor & building \\

\texttt{EVAL-226} & a dental room with dental chair and sink counter & health\_\allowbreak{}science & indoor & room \\

\texttt{EVAL-227} & airport dropoff road with the terminal canopy and pedestrian bridges & transport\_\allowbreak{}outdoor & outdoor & street \\

\texttt{EVAL-228} & a bus terminal and bench rows & transit & indoor & building \\

\texttt{EVAL-229} & model shelves and a drafting room & work\_\allowbreak{}education & indoor & room \\

\texttt{EVAL-230} & ferry passenger cabin with stairwell and seat rows & transit & indoor & building \\

\texttt{EVAL-231} & the open plan office with desk clusters & work\_\allowbreak{}education & indoor & room \\

\texttt{EVAL-232} & city bus stop and a route kiosk & transport\_\allowbreak{}outdoor & outdoor & street \\

\texttt{EVAL-233} & the wall panels of a control room & industrial & indoor & building \\

\texttt{EVAL-234} & the built in wardrobe of a bedroom & residential & indoor & room \\

\texttt{EVAL-235} & a central well and the rural village square & residential\_\allowbreak{}outdoor & outdoor & street \\

\texttt{EVAL-236} & central courtyard of a cloister interior & culture\_\allowbreak{}civic & indoor & building \\

\texttt{EVAL-237} & the open highway with distant hills and lane markings & transport\_\allowbreak{}outdoor & outdoor & street \\

\texttt{EVAL-238} & coastal boardwalk with the dune vegetation & coastal & outdoor & site \\

\texttt{EVAL-239} & surrounding buildings and stone paving of the city plaza & urban\_\allowbreak{}core & outdoor & street \\

\texttt{EVAL-240} & grassy path of a stone pasture & rural & outdoor & landscape \\

\texttt{EVAL-241} & reservoir overlook with an access path & parks\_\allowbreak{}water & outdoor & site \\

\texttt{EVAL-242} & cinema auditorium interior and stepped aisles & culture\_\allowbreak{}civic & indoor & building \\

\texttt{EVAL-243} & fabrication tables and steel beams of shipyard workshop interior & industrial & indoor & building \\

\texttt{EVAL-244} & the forest trail and tree canopy & forest & outdoor & landscape \\

\texttt{EVAL-245} & bakery interior with the display counter & food\_\allowbreak{}drink & indoor & room \\

\texttt{EVAL-246} & stone valley of rock pillars and sandy path & arid & outdoor & landscape \\

\texttt{EVAL-247} & historic railway depot and the brick station & heritage & outdoor & site \\

\texttt{EVAL-248} & steel trusses of the river bridge & transport\_\allowbreak{}outdoor & outdoor & street \\

\texttt{EVAL-249} & small tables and bay windows of the breakfast room & hospitality & indoor & room \\

\texttt{EVAL-250} & a department store with an escalator bank and display islands & retail & indoor & room \\

\texttt{EVAL-251} & layered cliffs of stone valley & arid & outdoor & landscape \\

\texttt{EVAL-252} & commercial kitchen with a tiled floor and stainless counters & food\_\allowbreak{}drink & indoor & room \\

\texttt{EVAL-253} & dirt roads and weathered buildings of the mining settlement & arid & outdoor & landscape \\

\texttt{EVAL-254} & the coastal path and the lighthouse grounds & coastal & outdoor & site \\

\texttt{EVAL-255} & the formal public garden with geometric paths & parks\_\allowbreak{}water & outdoor & site \\

\texttt{EVAL-256} & the alpine lakeshore with mountain backdrop and rocky shore & mountain & outdoor & landscape \\

\texttt{EVAL-257} & the town hall with ceremonial stairs and arched windows & culture\_\allowbreak{}civic & indoor & building \\

\texttt{EVAL-258} & overhead cranes and shipyard workshop & industrial & indoor & building \\

\texttt{EVAL-259} & an alpine lakeshore with rocky shore and footpath & mountain & outdoor & landscape \\

\texttt{EVAL-260} & surrounding buildings and city plaza & urban\_\allowbreak{}core & outdoor & street \\

\texttt{EVAL-261} & storage cabinets and lab benches of the science classroom & work\_\allowbreak{}education & indoor & room \\

\texttt{EVAL-262} & laboratory and work benches with fume hoods & health\_\allowbreak{}science & indoor & room \\

\texttt{EVAL-263} & redwood grove and fern understory & forest & outdoor & landscape \\

\texttt{EVAL-264} & the forest waterfall with rock basin and mossy ledges & forest & outdoor & landscape \\

\texttt{EVAL-265} & the housing district and sidewalk network & residential\_\allowbreak{}outdoor & outdoor & street \\

\texttt{EVAL-266} & hallway with multiple doorways and the stair landing & residential & indoor & room \\

\texttt{EVAL-267} & the rocky slopes and mountain pass & mountain & outdoor & landscape \\

\texttt{EVAL-268} & wood floor of the bedroom & residential & indoor & room \\

\texttt{EVAL-269} & the reservoir overlook and a guard railing & parks\_\allowbreak{}water & outdoor & site \\

\texttt{EVAL-270} & wood pews and central nave of a church interior & culture\_\allowbreak{}civic & indoor & building \\

\texttt{EVAL-271} & walton village hall school lane west yorkshire & culture\_\allowbreak{}civic & outdoor & building \\

\texttt{EVAL-272} & art museum & culture\_\allowbreak{}civic & indoor & building \\

\texttt{EVAL-273} & table tennis room & recreation & indoor & room \\

\texttt{EVAL-274} & chatsworth house & heritage & outdoor & site \\

\texttt{EVAL-275} & bell tower of vilnius cathedral & heritage & outdoor & building \\

\texttt{EVAL-276} & toronto first fire hall & heritage & outdoor & building \\

\texttt{EVAL-277} & knock community hall & culture\_\allowbreak{}civic & outdoor & building \\

\texttt{EVAL-278} & felton california bridge & transport\_\allowbreak{}outdoor & outdoor & site \\

\texttt{EVAL-279} & capitoline museum courtyard & culture\_\allowbreak{}civic & outdoor & site \\

\texttt{EVAL-280} & leake street graffiti tunnel & urban\_\allowbreak{}core & indoor & site \\

\texttt{EVAL-281} & st nicholas church in roxby & heritage & outdoor & building \\

\texttt{EVAL-282} & sunnyvale heritage park museum & heritage & outdoor & site \\

\texttt{EVAL-283} & a garden house in provence & residential\_\allowbreak{}outdoor & outdoor & site \\

\texttt{EVAL-284} & national maritime museum of korea & culture\_\allowbreak{}civic & outdoor & building \\

\texttt{EVAL-285} & english garden & parks\_\allowbreak{}water & outdoor & site \\

\texttt{EVAL-286} & norman constables house & heritage & outdoor & building \\

\texttt{EVAL-287} & christchurch castle & heritage & outdoor & site \\

\texttt{EVAL-288} & toronto roundhouse train yard & industrial & outdoor & site \\

\texttt{EVAL-289} & cochem imperial castle germany & heritage & outdoor & site \\

\texttt{EVAL-290} & dining room & food\_\allowbreak{}drink & indoor & room \\

\texttt{EVAL-291} & cactus garden & arid & outdoor & site \\

\texttt{EVAL-292} & a high-detail garage workshop interior & industrial & indoor & room \\

\texttt{EVAL-293} & grand courts art gallery of nsw & culture\_\allowbreak{}civic & indoor & building \\

\texttt{EVAL-294} & schiller lookout tower kryry czech republic & mountain & outdoor & site \\

\texttt{EVAL-295} & smederevo fortress serbia & heritage & outdoor & site \\

\texttt{EVAL-296} & avoncroft museum postmill & rural & outdoor & site \\

\texttt{EVAL-297} & country hotel le querce italy & hospitality & outdoor & building \\

\texttt{EVAL-298} & freilicht museum molfsee wind mill & rural & outdoor & site \\

\texttt{EVAL-299} & point lowly lighthouse south australia & coastal & outdoor & site \\

\texttt{EVAL-300} & regina coeli church & heritage & outdoor & building \\

\end{longtable}
\clearpage
\renewcommand{\arraystretch}{1.05}
\fussy
\normalsize

\bibliographystyle{plainnat}
\bibliography{references}

\end{document}